\pdfoutput=1
\documentclass[11pt]{article}

\usepackage{ACL2023}

\usepackage{times}
\usepackage{latexsym}

\usepackage[T1]{fontenc}
\usepackage[utf8]{inputenc}

\usepackage{microtype}

\usepackage{inconsolata}

\usepackage{algorithm}
\usepackage{algpseudocode}
\usepackage{booktabs}
\usepackage{multirow}
\usepackage{multicol}
\usepackage{graphicx}
\usepackage{amsmath}
\usepackage{caption}
\usepackage{colortbl}
\usepackage{arydshln}
\usepackage{xcolor}
\usepackage{tikz}
\usepackage{listings}
\usepackage{xcolor}
\definecolor{NavyBlue}{RGB}{35,67,121}
\definecolor{OliveGreen}{RGB}{77,135,56}

\algnewcommand\Output{\item[\textbf{Output:}]}

\title{Hierarchical Context Pruning: Optimizing Real-World Code Completion with Repository-Level Pretrained Code LLMs}

\author{
Lei Zhang\textsuperscript{1,2} \quad
Yunshui Li\textsuperscript{1,2} \quad
Jiaming Li\textsuperscript{1,2} \quad
Xiaobo Xia\textsuperscript{3} \quad
Jiaxi Yang\textsuperscript{1,2} \quad
Run Luo\textsuperscript{1,2} \quad
\\
\textbf{Minzheng Wang}\textsuperscript{2,5}  \quad
\textbf{Longze Chen}\textsuperscript{1,2}  \quad
\textbf{Junhao Liu}\textsuperscript{4} \quad
\textbf{Min Yang}\textsuperscript{1,2}\footnotemark[2] \\
\textsuperscript{1}Shenzhen Institute of Advanced Technology, Chinese Academy of Sciences\\
\textsuperscript{2}University of Chinese Academy of Sciences\\
\textsuperscript{3}The University of Sydney \quad
\textsuperscript{4}University of California, Irvine \\
\textsuperscript{5}MAIS, Institute of Automation, Chinese Academy of Sciences\\
\texttt{\{lei.zhang2, min.yang\}@siat.ac.cn}
}

\begin{document}

\maketitle

\renewcommand{\thefootnote}{\fnsymbol{footnote}}
\footnotetext[2]{Min Yang is the corresponding author.}

\renewcommand{\thefootnote}{\arabic{footnote}}

\begin{abstract}
  Some recently developed code large language models (Code LLMs) have been pretrained on repository-level code data (Repo-Code LLMs), enabling these models to recognize repository structures and utilize cross-file information for code completion.
  However, in real-world development scenarios, simply concatenating the entire code repository often exceeds the context window limits of these Repo-Code LLMs, leading to significant performance degradation.
  In this study, we conducted extensive preliminary experiments and analyses on six Repo-Code LLMs.
  The results indicate that maintaining the topological dependencies of files and increasing the code file content in the completion prompts can improve completion accuracy;
  pruning the specific implementations of functions in all dependent files does not significantly reduce the accuracy of completions.
  Based on these findings, we proposed a strategy named \textbf{Hierarchical Context Pruning (HCP)} to construct completion prompts with high informational code content.
  The \textbf{HCP} models the code repository at the function level, maintaining the topological dependencies between code files while removing a large amount of irrelevant code content, significantly reduces the input length for repository-level code completion.
  We applied the \textbf{HCP} strategy in experiments with six Repo-Code LLMs, and the results demonstrate that our proposed method can significantly enhance completion accuracy while substantially reducing the length of input.
  Our code and data are available at \url{https://github.com/Hambaobao/HCP-Coder}.
\end{abstract}

\section{Introduction}
Code completion tools based on code large language models \citep{chen2021evaluating,nijkamp2023codegen,li2023starcoder,fried2023incoder,allal2023santacoder}, such as \emph{GitHub Copilot}\footnote{\url{https://github.com/features/copilot}}, have been widely adopted in daily development practices and have significantly enhanced the productivity of developers.
As research \citep{bavarian2022efficient,sun2024survey} on code large language models (Code LLMs) continues to evolve, some recently developed Code LLMs \citep{guo2024deepseekcoder,lozhkov2024starcoder,codegemma_2024} have been trained on repository-level code data (Repo-Code LLMs) to overcome the limitations of previous models trained on file-level data, which struggled to recognize repository structures and integrate code across multiple files for completion tasks.
However, in real-world development scenarios, simply concatenating the entire code repository often exceeds the context window size of these Repo-Code LLMs, leading to significant performance degradation and increased inference latency.
How to effectively utilize the capabilities of these Repo-Code LLMs to integrate cross-file information and construct high-quality completion prompts within the model’s context window limits remains an area for further exploration.

In this study, we initially evaluated six Repo-Code LLMs on the CrossCodeEval \citep{ding2023crosscodeeval} benchmark and conducted a detailed analysis of completion errors (Appendix \ref{sec:error_analysis}).
The errors identified were categorized into eight distinct classes (Section \ref{sec:error_analysis_main}).
Subsequently, considering the characteristics of the decoder architecture in Code LLMs, we analyzed the impact of topological dependencies among code files on completion accuracy (Section \ref{sec:dependency_analysis}).
We found that maintaining the dependencies between code files and including more file information leads to higher accuracy.
Additionally, we conducted experiments to analyze the impact of content from files at different dependency levels on completion accuracy (Section \ref{sec:cross_file_content_analysis}).
We discovered that even pruning away the specific implementations of functions in all dependent files does not significantly reduce the accuracy of completions.
Based on the results of these preliminary experiments, we proposed a strategy named \textbf{Hierarchical Context Pruning (HCP)} to construct high-quality completion prompts.
The \textbf{HCP} models the code repository at the function level, retaining the topological dependencies between files while eliminating a large amount of irrelevant code content.
In our experiments, the \textbf{HCP} successfully reduced the input from over 50,000 tokens to approximately 8,000 tokens, and significantly enhanced the accuracy of completions.

In summary, our contributions are threefold:
\begin{itemize}
  \item We conducted experiments on six Repo-Code LLMs and found that: maintaining the topological dependencies of files and increasing the content of code files in the completion prompts can enhance completion accuracy; pruning the specific implementations of functions in all dependent files does not significantly reduce the accuracy of completions.
  \item Based on the results of preliminary experiments, we proposed a strategy named \textbf{Hierarchical Context Pruning (HCP)} for constructing high-quality completion prompts, which models the code repository at the function level, retaining the topological dependencies between files while eliminating a large amount of irrelevant code content.
  \item We applied the \textbf{HCP} strategy in experiments with six Repo-Code LLMs, and the results demonstrate that our proposed method can significantly enhance completion accuracy while substantially reducing the length of input.
\end{itemize}

\section{Related Work}
\subsection{Code Large Language Models}
\subsubsection{Infilling Code LLMs}
Infilling scenarios constitute the majority of code completion tasks in the real world.
\citet{bavarian2022efficient} demonstrates that pre-training Code LLMs with a certain proportion of fill-in-the-middle format code data can enable the Code LLMs to fill in middle code based on the surrounding context, without compromising their original left-to-right generation performance.
Based on the findings of \citet{bavarian2022efficient}, many subsequent Code LLMs \citep{fried2023incoder,allal2023santacoder,nijkamp2023codegen2,li2023starcoder,roziere2024code,guo2024deepseekcoder,pinnaparaju2024stable,lozhkov2024starcoder} have emerged with the capability to perform infilling.

\subsubsection{Instruction Code LLMs}
Pretrained Code LLMs are traditionally used only for continuation tasks such as code completion.
Inspired by works on instruction tuning large language models \citep{ouyang2022training,li2024oneshot}, many studies \citep{wang2023far,luo2023wizardcoder,muennighoff2024octopack,xu2023lemur,wang2024dolphcoder,zheng2024opencodeinterpreter} have attempted to finetune Code LLMs using code instruction data.
This finetuning unlocks the potential of Code LLMs, enabling them to perform more complex coding tasks based on user instructions.

\begin{figure*}[h]
  \centering
  \includegraphics[width=0.98\textwidth]{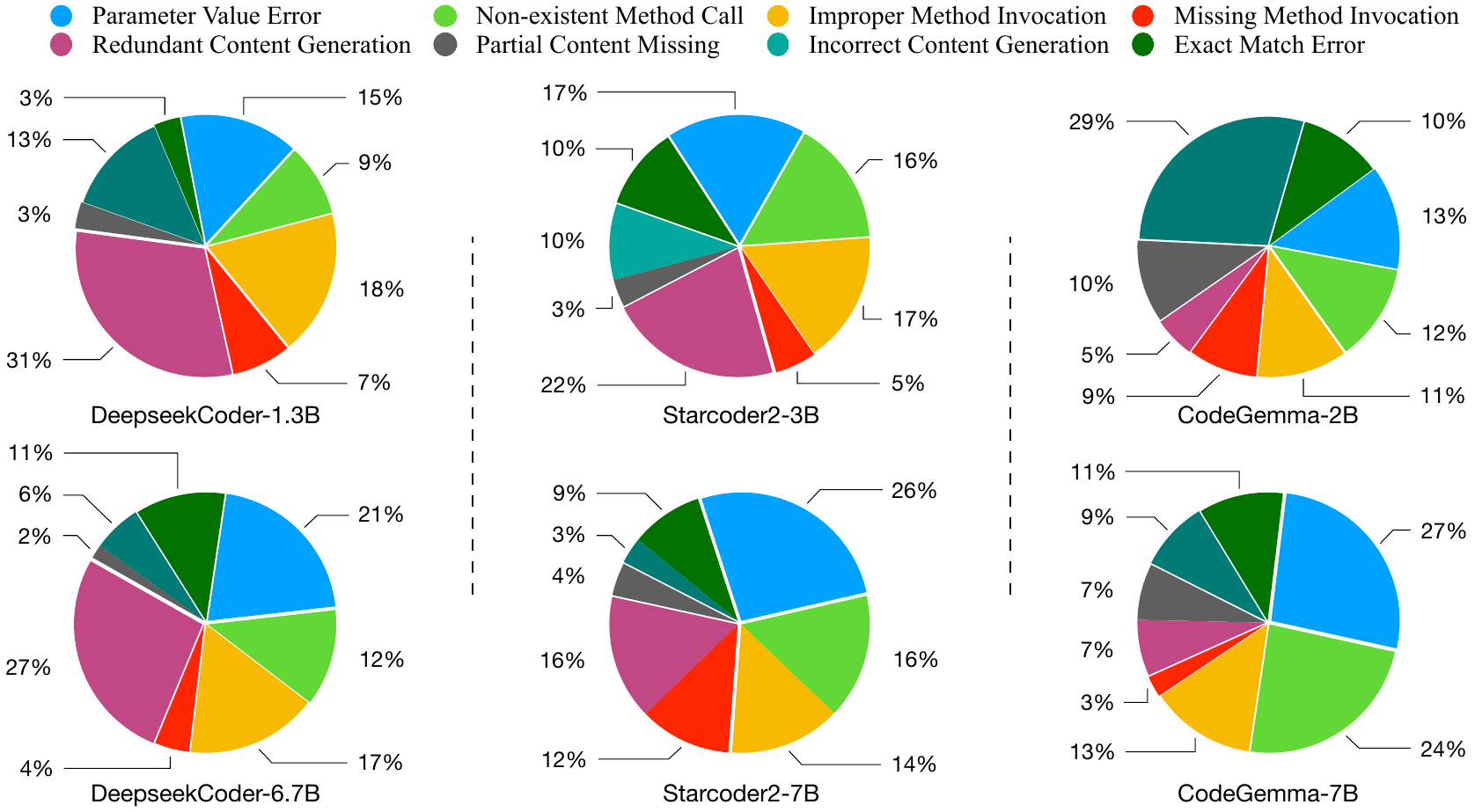}
  \caption{The error class distribution of the completion results of the DeepseekCoder, Starcoder2 and CodeGemma models on the CrossCodeEval: Python benchmark.}
  \label{fig:case_study_dist}
\end{figure*}

\subsection{Code Benchmarks}
\subsubsection{Code Completion Benchmarks}
HumanEval \citep{chen2021evaluating} consists of 164 manually crafted Python code problems, with an average of 7.7 tests each test case.
MBPP \citep{austin2021program} is designed for individuals with entry-level programming skills. It comprises 974 concise Python functions, each with an accompanying description in English, a specified function signature, and three manually crafted test cases for verification.
MultiPL-E \citep{cassano2022multiple} introduces itself as a novel benchmarking framework designed for multilingual contexts, building upon HumanEval \citep{chen2021evaluating} and MBPP \citep{austin2021program}.
APPS \citep{hendrycks2021measuring} is a benchmark including 10K less-restricted problems for code generation.
CodeContests \citep{Li-2022} is a dataset specifically for competitive programming problems.

\subsubsection{Infilling Code Benchmarks}
\citet{fried2023incoder} constructed \textit{single-line} and \textit{multi-line} infilling completion tasks based on HumanEval, and \citet{bavarian2022efficient} expanded upon it to create \textit{randomspan} infilling completion tasks, ultimately resulting in the current HumanEval-Infilling benchmark.
\citet{allal2023santacoder} created an Infilling benchmark that includes languages from \texttt{Java}, \texttt{JavaScript}, and \texttt{Python 3}, utilizing a line exactly match method for evaluation.
\citet{lai2022ds1000} presents a benchmark for evaluating the performance of Code LLMs in completing tasks related to Python scientific computing libraries, encompassing both regular completion and insertion (infilling) tasks.

\subsection{Repo-level Code Completion}
Some benchmarks for repository-level code completion have been proposed to evaluate the performance of code models in real-world completion tasks, such as CrossCodeEval \citep{ding2023crosscodeeval}, Repo-Bench \citep{liu2023repobench}, CoderEval \citep{Zhang_2024}, and EvoCodeBench \citep{li2024evocodebench}.
A lot of studies \citep{shrivastava2023repositorylevel,zhang2023repocoder,bi2024iterative,phan2024repohyper,liang2024repofuse} have focused on improving the accuracy of repository-level code completion tasks.
However, most of these studies overlook the unique aspects of their Fill-in-the-Middle (FIM) capacities.
Furthermore, despite the recent development of repository-level pretrained Code LLMs designed to process large-scale repository data, research on these models remains relatively limited.

\section{Experiments Setup}
\subsection{Dataset \& Evaluation Metrics}
To assess the code completion performance of Code LLMs in real development scenarios, we utilized CrossCodeEval \citep{ding2023crosscodeeval} as the evaluation dataset.
The CrossCodeEval \citep{ding2023crosscodeeval} benchmark provides test cases that require the use of cross-file code information for completion.
Without loss of generality, in this study, we have chosen \texttt{Python} language as the primary language for our research.

We used the original data from CrossCodeEval, retaining the original repository structure.
For each test case, we first identified the file for completion and the cursor's position (the line and column where the completion occurs).
We then removed the code after the cursor in that line to form authentic completion test cases.
Ultimately, we obtained 2,655 real-world completion tests.
Following the CrossCodeEval evaluation protocol, we evaluated the completion results using two metrics: \emph{Exact Match} (EM) and \emph{Edit Similarity} (ES).

\subsection{Models \& Prompt Templates}
The code large language models pretrained with repository-level code data include specific tokens used to describe the repository structure in the prompt.
Table \ref{tab:special_tokens} in appendix displays the special tokens used by DeepseekCoder, Starcoder2 and CodeGemma.
The specific prompt templates used by DeepseekCoder, Starcoder2 and CodeGemma are shown in Table \ref{tab:prompt_template}.

\subsection{Hardware \& Hyperparameters}
All the expiriments were conducted on NVIDIA A100 GPUs.
We employ greedy decoding strategy for all the models, and set \texttt{max\_new\_tokens} to $32$.
The \texttt{model\_max\_length} of DeepseekCoder, Starcoder2 and CodeGemma is set to $16,352$, $16,352$ and $8,160$, respectively.
All the prompts longer than the \texttt{model\_max\_length} are truncated from the left.

\begin{table*}[h]
  \setlength\dashlinedash{1pt}
  \setlength\dashlinegap{1pt}
  \centering
  \resizebox{\textwidth}{!}{
    \begin{tabular}{l r r r r r r r r r r r r}
      \toprule
      \multirow{3}{*}{\bf XF-Context}                                                                                       & \multicolumn{12}{c}{\textbf{Baseline Evaluation}}                                                                                                                                                                                                                                                                                    \\
      \cmidrule(lr){2-13}                                                                                                   & \multicolumn{2}{c}{\textbf{DScoder-1.3B }}        & \multicolumn{2}{c}{\textbf{DScoder-6.7B }} & \multicolumn{2}{c}{\textbf{Starcoder2-3B }} & \multicolumn{2}{c}{\textbf{Starcoder2-7B }} & \multicolumn{2}{c}{\textbf{CodeGemma-2B }} & \multicolumn{2}{c}{\textbf{CodeGemma-7B }}                                                 \\
      \cmidrule(lr){2-3} \cmidrule(lr){4-5} \cmidrule(lr){6-7} \cmidrule(lr){8-9} \cmidrule(lr){10-11} \cmidrule(lr){12-13} & {EM}                                              & {ES}                                       & {EM}                                        & {ES}                                        & {EM}                                       & {ES}                                       & {EM}  & {ES}  & {EM}  & {ES}  & {EM}  & {ES}  \\
      \midrule
      \textbf{Infile-Only}                                                                                                  & 16.72                                             & 56.58                                      & 28.14                                       & 68.36                                       & 21.92                                      & 61.49                                      & 22.98 & 63.58 & 20.64 & 56.26 & 30.58 & 70.36 \\
      \midrule
      \textbf{RAG-BM25}                                                                                                     & 17.28                                             & 58.18                                      & 32.65                                       & 71.78                                       & 24.45                                      & 63.84                                      & 26.26 & 65.32 & 22.89 & 57.73 & 32.89 & 70.81 \\
      % \; + \textit{reference}                                                                                               & \textit{19.40}                                    & \textit{59.08}                             & \textit{36.29}                              & \textit{72.99}                              & \textit{28.74}                             & \textit{66.01}                             & \textit{29.68} & \textit{67.01} & \textit{24.84} & \textit{58.32} & \textit{39.02} & \textit{75.16} \\
      \midrule
      \textbf{Random-All}                                                                                                   & 6.18                                              & 46.19                                      & 33.94                                       & 70.98                                       & 28.32                                      & 66.87                                      & 31.45 & 69.09 & 26.93 & 62.13 & 36.69 & 74.42 \\
      \bottomrule
    \end{tabular}
  }
  \caption{
    The completion results of the baseline methods.
    \textbf{EM} denotes Exact Match, and \textbf{ES} denotes Edit Similarity.
  }
  \label{tab:cross_file_result}
\end{table*}

\begin{table*}[h]
  \setlength\dashlinedash{1pt}
  \setlength\dashlinegap{1pt}
  \centering
  \resizebox{\textwidth}{!}{
    \begin{tabular}{l r r r r r r r r r r r r}
      \toprule
      \multirow{3}{*}{\bf XF-Context}                                                                                       & \multicolumn{12}{c}{\textbf{Topoligical Dependency Analysis}}                                                                                                                                                                                                                                                                                                                                          \\
      \cmidrule(lr){2-13}                                                                                                   & \multicolumn{2}{c}{\textbf{DScoder-1.3B }}                    & \multicolumn{2}{c}{\textbf{DScoder-6.7B }} & \multicolumn{2}{c}{\textbf{Starcoder2-3B }} & \multicolumn{2}{c}{\textbf{Starcoder2-7B }} & \multicolumn{2}{c}{\textbf{CodeGemma-2B }} & \multicolumn{2}{c}{\textbf{CodeGemma-7B }}                                                                                                       \\
      \cmidrule(lr){2-3} \cmidrule(lr){4-5} \cmidrule(lr){6-7} \cmidrule(lr){8-9} \cmidrule(lr){10-11} \cmidrule(lr){12-13} & {EM}                                                          & {ES}                                       & {EM}                                        & {ES}                                        & {EM}                                       & {ES}                                       & {EM}           & {ES}           & {EM}           & {ES}           & {EM}           & {ES}           \\
      \midrule
      \textbf{D-Level: 1}                                                                                                   & \bf 15.44                                                     & \bf 55.03                                  & 33.03                                       & 70.77                                       & 26.18                                      & 64.15                                      & 28.51          & 66.91          & 24.37          & 58.79          & 34.65          & 73.01          \\
      \cmidrule(lr){1-1}\cmidrule(lr){2-3} \cmidrule(lr){4-5} \cmidrule(lr){6-7} \cmidrule(lr){8-9} \cmidrule(lr){10-11} \cmidrule(lr){12-13}
      \textbf{D-Level: 2}                                                                                                   & 13.63                                                         & 53.45                                      & 33.56                                       & 70.74                                       & 26.70                                      & 64.58                                      & 29.45          & 67.03          & 25.31          & 59.27          & 35.67          & 73.26          \\
      \cmidrule(lr){1-1}\cmidrule(lr){2-3} \cmidrule(lr){4-5} \cmidrule(lr){6-7} \cmidrule(lr){8-9} \cmidrule(lr){10-11} \cmidrule(lr){12-13}
      \textbf{D-Level: 3}                                                                                                   & 13.26                                                         & 53.17                                      & 33.07                                       & 70.51                                       & 26.82                                      & 64.56                                      & 29.23          & 67.01          & 25.35          & 59.30          & 35.93          & 73.34          \\
      \cmidrule(lr){1-1}\cmidrule(lr){2-3} \cmidrule(lr){4-5} \cmidrule(lr){6-7} \cmidrule(lr){8-9} \cmidrule(lr){10-11} \cmidrule(lr){12-13}
      \textbf{D-Level: 4}                                                                                                   & 13.37                                                         & 53.20                                      & 33.22                                       & 70.57                                       & 26.59                                      & 64.46                                      & 29.53          & 67.07          & 25.54          & 59.42          & 36.12          & 73.54          \\
      \cmidrule(lr){1-1}\cmidrule(lr){2-3} \cmidrule(lr){4-5} \cmidrule(lr){6-7} \cmidrule(lr){8-9} \cmidrule(lr){10-11} \cmidrule(lr){12-13}
      \textbf{D-Level: $\infty$}                                                                                            & 5.76                                                          & 46.22                                      & \textbf{35.29}                              & \textbf{71.51}                              & \textbf{30.43}                             & \textbf{67.34}                             & \textbf{33.03} & \textbf{69.57} & \textbf{29.08} & \textbf{62.91} & \textbf{39.32} & \textbf{75.35} \\
      \bottomrule
    \end{tabular}
  }
  \caption{
    Comparison of completion results using different context dependency levels across 6 models.
    All the prompts is truncated to the max context window of the Code LLMs from the left.
    \emph{$\infty$} denotes the prompt including all files in the repository.
  }
  \label{tab:dependency_analysis}
\end{table*}

\section{Preliminary Studies}
\subsection{Baseline Evaluation}
\subsubsection{Infile Only}
We initially evaluated the model's completion ability using only information from the current file, with results presented in Table \ref{tab:cross_file_result} under the \textit{Infile-Only} row.
The completion results are less than satisfactory.
Even the best-performing model achieved an accuracy of only about 30\%.

\begin{figure}[h]
  \centering
  \includegraphics[width=\columnwidth]{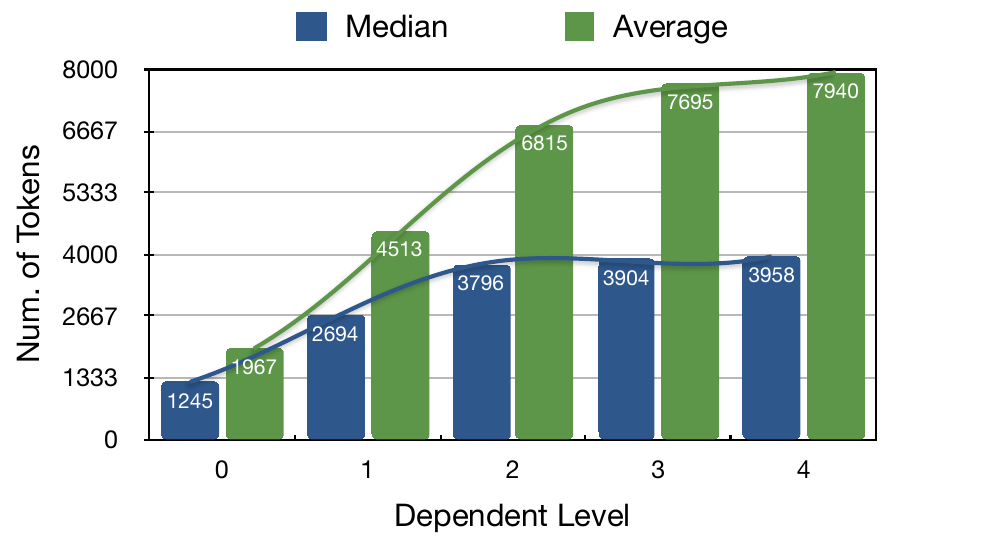}
  \caption{The distribution of tokenized prompt lengths in the CrossCodeEval benchmark. The x-aixs represents the dependent level, and the y-axis represents the number of tokens.
  }
  \label{fig:tokenized_prompt_length_distribution_single}
\end{figure}

\subsubsection{RAG-BM25}
We subsequently evaluated the effect of using Retrieval Augmented Generation (RAG) method to retrieve relevant code snippets to assist with completion.
Following the setup of CrossCodeEval, we chunk the repository code into units of 10 lines, and use BM25 as similarity metric for retrieving relevant code snippets.
We select the top-5 relevant snippets as cross-file information, which are placed at the beginning of the prompt to assist with code generation.
The results are shown in Table \ref{tab:cross_file_result} under the \textit{RAG-BM25} row.

\begin{table*}[h]
  \setlength\dashlinedash{1pt}
  \setlength\dashlinegap{1pt}
  \centering
  \resizebox{\textwidth}{!}{
    \begin{tabular}{l r r r r r r r r r r r r}
      \toprule
      \multirow{3}{*}{\bf XF-Context}                                                                                       & \multicolumn{12}{c}{\textbf{Cross-File Content Analysis}}                                                                                                                                                                                                                                                                                                            \\
      \cmidrule(lr){2-13}                                                                                                   & \multicolumn{2}{c}{\textbf{DScoder-1.3B }}                & \multicolumn{2}{c}{\textbf{DScoder-6.7B }} & \multicolumn{2}{c}{\textbf{Starcoder2-3B }} & \multicolumn{2}{c}{\textbf{Starcoder2-7B }} & \multicolumn{2}{c}{\textbf{CodeGemma-2B }} & \multicolumn{2}{c}{\textbf{CodeGemma-7B }}                                                                         \\
      \cmidrule(lr){2-3} \cmidrule(lr){4-5} \cmidrule(lr){6-7} \cmidrule(lr){8-9} \cmidrule(lr){10-11} \cmidrule(lr){12-13} & {EM}                                                      & {ES}                                       & {EM}                                        & {ES}                                        & {EM}                                       & {ES}                                       & {EM}      & {ES}      & {EM}      & {ES}      & {EM}      & {ES}      \\
      \midrule
      \textbf{P-Level: 0}                                                                                                   & 6.18                                                      & 46.19                                      & 33.94                                       & 70.98                                       & 28.32                                      & 66.87                                      & 31.45     & 69.09     & 26.93     & 62.13     & 36.69     & 74.42     \\
      \midrule
      \textbf{P-Level: 1}                                                                                                   & 6.55                                                      & 46.58                                      & 36.20                                       & 71.90                                       & \bf 30.73                                  & \bf 67.97                                  & \bf 34.43 & \bf 70.65 & \bf 29.30 & \bf 63.46 & \bf 39.55 & \bf 75.70 \\
      \midrule
      \textbf{P-Level: 2}                                                                                                   & \bf 9.83                                                  & \bf 49.63                                  & 34.73                                       & 70.89                                       & 30.02                                      & 66.41                                      & 31.26     & 68.24     & 27.34     & 61.13     & 38.31     & 74.32     \\
      \cmidrule(lr){2-3} \cmidrule(lr){4-5} \cmidrule(lr){6-7} \cmidrule(lr){8-9} \cmidrule(lr){10-11} \cmidrule(lr){12-13}
      \; + \textit{D-level:1}                                                                                               & 9.45                                                      & 49.44                                      & \bf 36.87                                   & \bf 72.14                                   & 29.91                                      & 66.96                                      & 32.62     & 69.11     & 28.93     & 62.03     & 39.17     & 75.16     \\
      \cmidrule(lr){2-3} \cmidrule(lr){4-5} \cmidrule(lr){6-7} \cmidrule(lr){8-9} \cmidrule(lr){10-11} \cmidrule(lr){12-13}
      \; + \textit{D-level:2}                                                                                               & 8.70                                                      & 48.61                                      & 36.38                                       & 71.66                                       & 29.64                                      & 66.99                                      & 32.96     & 69.13     & 28.44     & 61.76     & 39.06     & 74.91     \\
      \bottomrule
    \end{tabular}
  }
  \caption{
    The results of completion using cross-file information with different pruning levels.
    \emph{+ D-level:x} denotes the model uses the cross-file information with dependency level x.
  }
  \label{tab:sparse_cross_file_result}
\end{table*}

\subsubsection{Randomly Concatenating All Files}
Additionally, we concatenated all repository code files randomly according to the pre-trained formats of various Repo-Code LLMs to create completion prompts, which were then input into the models for completion.
The evaluation results are shown in Table \ref{tab:cross_file_result} under the \textit{Random-All} row.
We observed that supplying the model with more information from the repository's code led to superior performance compared to RAG.
However, the input length of the model is limited by its context window, thereby transforming this scenario into a constrained optimization problem.
The constrained optimization goal is expressed as follows:
\begin{equation}
  \begin{aligned}
     & \max_{\mathcal{P}} \text{Quality}(\mathcal{P}) \quad\text{s.t.} \quad \text{Length}(\mathcal{P}) \leq L
  \end{aligned}
\end{equation}
where
\( \mathcal{P} \) represents the coonstructed prompt, \( \text{Quality}(\mathcal{P}) \) represents the quality of the coonstructed prompt, \( \text{Length}(\mathcal{P}) \) represents the length of the constructed prompt, and \( L \) represents context window size of the model.

\subsection{Completion Error Analysis}
\label{sec:error_analysis_main}
To further investigate the issues of repository-level pre-trained Code LLMs in real-world completion tasks, we sampled 200 error examples from each model's \emph{Random-All} evaluation results for error analysis.
Ultimately, we categorized the issues present in these models into eight classes:
\emph{Parameter Value Error}, \emph{Non-existent Method Call}, \emph{Improper Method Invocation}, \emph{Missing Method Invocation}, \emph{Redundant Content Generation}, \emph{Partial Content Missing}, \emph{Incorrect Content Generation}, and \emph{Exact Match Error}.
Figure \ref{fig:case_study_dist} shows the error distribution statistics for six Repo-Code LLMs.
In the appendix \ref{sec:error_analysis}, we provide examples of each type of error along with corresponding error analysis.

\begin{figure*}[h]
  \centering
  \includegraphics[width=\textwidth]{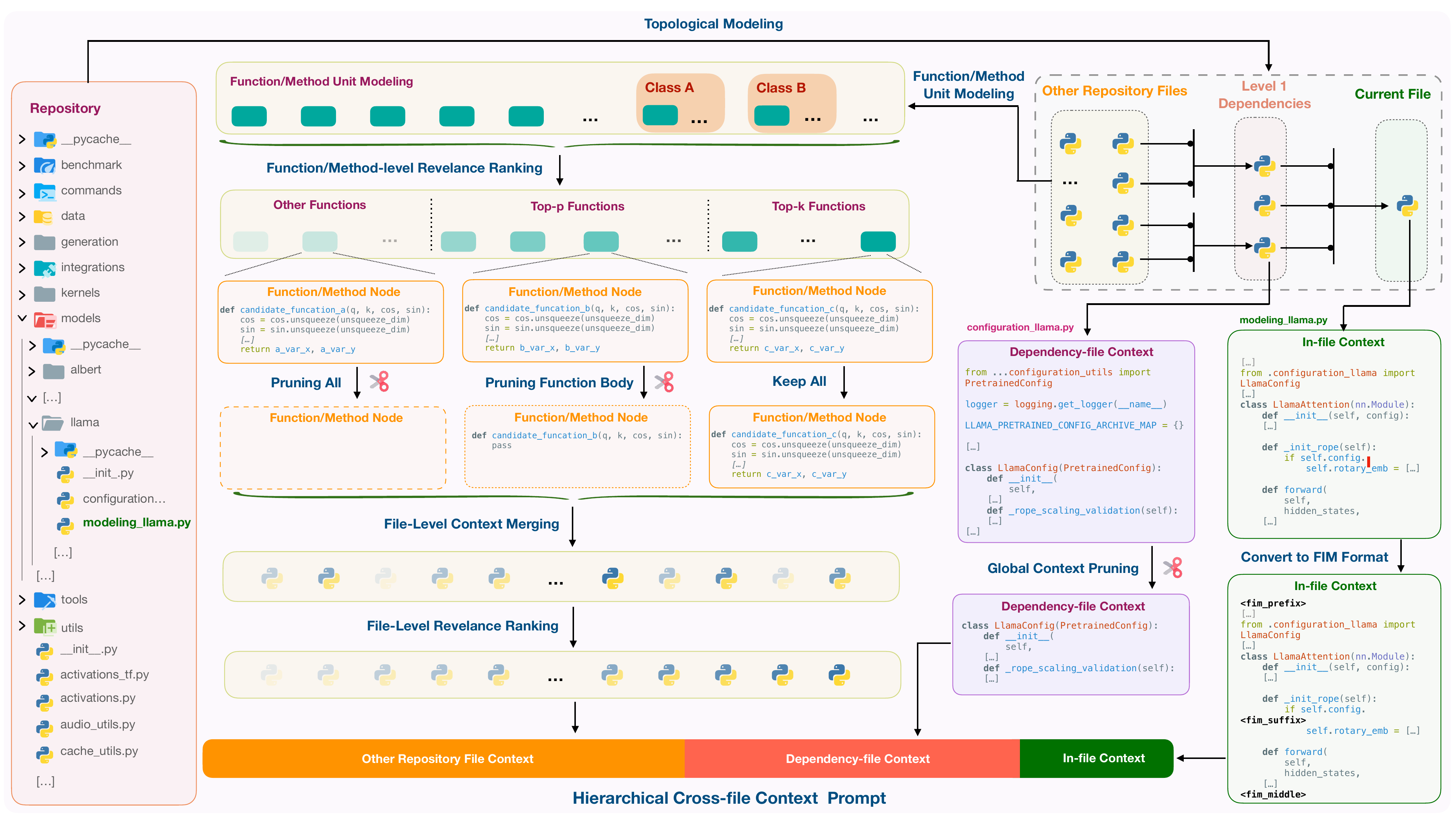}
  \caption{
    The framework of hierarchical context pruning for improving the performance of code large language models in real-world code completion tasks.
  }
  \label{fig:methedology}
\end{figure*}

\subsection{Topological Dependency Analysis}
\label{sec:dependency_analysis}
\paragraph{Definition 1.} (Dependency Level)
Let $F$ denote a set of files in a code repository, and let $f \in F$ represent a specific file. We define the dependency levels as follows:
\begin{equation}
  \begin{aligned}
     & I(f)       = \{g \mid g \text{ is imported by } f\} \\
     & D_0(f)     = \{f\}                                  \\
     & D_{i+1}(f) = D_i(f) \cup I(D_i(f))                  \\
    %  & D_{\infty}(f) =  D_4(f) \cup \{F \setminus D_4(f)\} \\
  \end{aligned}
  \label{eq:dependency_levels}
\end{equation}

We first identified the file requiring completion, then extracted all the import statements from the file with \emph{Tree-Sitter}\footnote{\url{https://tree-sitter.github.io/tree-sitter}}, and used a breadth-first search (BFS) method to progressively add dependent files.
Algorithm \ref{alg:dependency-modeling} in appendix shows our specific dependency modeling process.

Figure \ref{fig:tokenized_prompt_length_distribution} illustrates the growth in the number of dependent files (calculated by the length of the tokenized prompt) as the number of dependency layers increases.
We used median and average as statistical measures and found that in the vast majority of cases, the number of dependent files for a single file increases slowly after reaching four layers of dependencies.
This suggests that using four layers of dependencies is sufficient to cover most scenarios.
We further define:
\begin{equation}
  \begin{aligned}
     & D_{\infty}(f) =  D_4(f) \cup \{F \setminus D_4(f)\}
  \end{aligned}
  \label{eq:dependency_infinite}
\end{equation}
to represent the prompt including all files in the repository.

In Table \ref{tab:dependency_analysis}, the D-level rows show the results of completion using cross-file information with different dependency levels.
The results indicate that although the maximum dependency depth of most files reaches 4 levels, only the information provided by $D_1(f)$ files is the most useful.
Furthermore, the effectiveness of using $D_{\infty}(f)$ surpasses that of Random-All, indicating that besides $D_1(f)$ files, there are many other useful files within the repository.

\subsection{Cross-File Content Analysis}
\label{sec:cross_file_content_analysis}
\paragraph{Definition 2.} (Pruning Level)
We define the pruning levels into three categories:
\begin{itemize}
  \item \textbf{P-Level 0}: No pruning is applied to the file content.
  \item \textbf{P-Level 1}: All global context content is removed from the file.
  \item \textbf{P-Level 2}: All global context content, function bodies and class method bodies are removed from the file.
\end{itemize}

Table \ref{tab:sparse_cross_file_result} presents the results of completion using cross-file information with different pruning levels.
We can see that the results of \emph{P-level:1} outperform those of \emph{P-level:0}, indicating that the Global Context information from cross-file content has minimal impact on the completion of the current file.
Additionally, the results of \emph{P-level:2} are only slightly worse than those of $D_{\infty}(f)$, and when combined with the information from $D_1(f)$, they are almost equivalent to the results of $D_{\infty}(f)$.
This suggests that the specific implementations of most cross-file functions have minimal impact on the completion of the current file, and retaining only the function header information is sufficient.

\section{Hierarchical Context Pruning}
Based on the analysis results concerning the dependencies and content of the files, we attempt to construct a hierarchical context prompt based on the importance and relevance of the repository content.
This approach aims to enhance the accuracy of code completion models while effectively reducing the length of the context.
Figure \ref{fig:methedology} shows the specific process for constructing a hierarchical context prompt.

\subsection{Fine-grained Repository Modeling}
In order to precisely control the content within the code repository, we employ \emph{Tree-Sitter} to parse the files within the repository.
We model the content using three types of nodes:
\begin{itemize}
  \item \textbf{Function Node}: Represents a function or a class method within a code file.
  \item \textbf{Class Node}: Represents a class in a code file, consisting of the class's name, attributes, and Function Nodes.
  \item \textbf{File Node}: Represents a code file, comprising Nodes that represent the functions and classes within the file, along with global context information.
\end{itemize}

\subsection{Hierarchical Context}
As shown in the top right of Figure \ref{fig:methedology}, following the settings in Section \ref{sec:dependency_analysis}, we conduct a dependency analysis on the files in the repository.
We perform a topological sort based on the dependency relationships, centering around the file currently being completed.
According to the experimental results in Section \ref{sec:dependency_analysis}, only files at dependency level 1 significantly enhance completion accuracy.
Therefore, we select files designated as $D_1(f)$ to serve as dependency files.
Ultimately, the files in the repository are categorized into three types: \emph{current file}, \emph{dependency files}, and \emph{other files}.
We will apply different strategies to optimize each type of file.

\begin{table*}[h]
  \setlength\dashlinedash{1pt}
  \setlength\dashlinegap{1pt}
  \centering
  \resizebox{\textwidth}{!}{
    \begin{tabular}{l r r r r r r r r r r r r}
      \toprule
      \multirow{3}{*}{\bf XF-Context}                                                                                       & \multicolumn{12}{c}{\textbf{Hierarchical Context Pruning (Top-p: 1.0)}}                                                                                                                                                                                                                                                                                                            \\
      \cmidrule(lr){2-13}                                                                                                   & \multicolumn{2}{c}{\textbf{DScoder-1.3B }}                              & \multicolumn{2}{c}{\textbf{DScoder-6.7B }} & \multicolumn{2}{c}{\textbf{Starcoder2-3B }} & \multicolumn{2}{c}{\textbf{Starcoder2-7B }} & \multicolumn{2}{c}{\textbf{CodeGemma-2B }} & \multicolumn{2}{c}{\textbf{CodeGemma-7B }}                                                                         \\
      \cmidrule(lr){2-3} \cmidrule(lr){4-5} \cmidrule(lr){6-7} \cmidrule(lr){8-9} \cmidrule(lr){10-11} \cmidrule(lr){12-13} & {EM}                                                                    & {ES}                                       & {EM}                                        & {ES}                                        & {EM}                                       & {ES}                                       & {EM}      & {ES}      & {EM}      & {ES}      & {EM}      & {ES}      \\
      \midrule
      \textbf{Random-All}                                                                                                   & 6.18                                                                    & 46.19                                      & 33.94                                       & 70.98                                       & 28.32                                      & 66.87                                      & 31.45     & 69.09     & 26.93     & 62.13     & 36.69     & 74.42     \\
      \midrule
      \textbf{Top-k: 0}                                                                                                     & 9.45                                                                    & 49.44                                      & 36.87                                       & 72.14                                       & 29.91                                      & 66.96                                      & 32.62     & 69.11     & 28.93     & 62.03     & 39.17     & 75.16     \\
      \cmidrule(lr){2-3} \cmidrule(lr){4-5} \cmidrule(lr){6-7} \cmidrule(lr){8-9} \cmidrule(lr){10-11} \cmidrule(lr){12-13}
      \textbf{Top-k: 5}                                                                                                     & 9.64                                                                    & 49.78                                      & 39.74                                       & 73.90                                       & 32.68                                      & 69.05                                      & 35.76     & 71.41     & 31.26     & 63.74     & 42.44     & 76.95     \\
      \cmidrule(lr){2-3} \cmidrule(lr){4-5} \cmidrule(lr){6-7} \cmidrule(lr){8-9} \cmidrule(lr){10-11} \cmidrule(lr){12-13}
      \textbf{Top-k: 10}                                                                                                    & \bf 9.91                                                                & \bf 49.85                                  & \bf 40.30                                   & \bf 74.56                                   & \bf 34.15                                  & \bf 69.37                                  & \bf 36.47 & \bf 71.50 & \bf 31.82 & \bf 64.34 & \bf 42.63 & \bf 77.35 \\
      \bottomrule
    \end{tabular}
  }
  \caption{
    The results of completion using hierarchical context pruning with different top-k values.
  }
  \label{tab:hierarchical_topk_context_result_short}
\end{table*}

\begin{table*}[h]
  \setlength\dashlinedash{1pt}
  \setlength\dashlinegap{1pt}
  \centering
  \resizebox{\textwidth}{!}{
    \begin{tabular}{l r r r r r r r r r r r r}
      \toprule
      \multirow{3}{*}{\bf XF-Context}                                                                                       & \multicolumn{12}{c}{\textbf{Hierarchical Context Pruning (Top-k: 5)}}                                                                                                                                                                                                                                                                                                            \\
      \cmidrule(lr){2-13}                                                                                                   & \multicolumn{2}{c}{\textbf{DScoder-1.3B }}                            & \multicolumn{2}{c}{\textbf{DScoder-6.7B }} & \multicolumn{2}{c}{\textbf{Starcoder2-3B }} & \multicolumn{2}{c}{\textbf{Starcoder2-7B }} & \multicolumn{2}{c}{\textbf{CodeGemma-2B }} & \multicolumn{2}{c}{\textbf{CodeGemma-7B }}                                                                         \\
      \cmidrule(lr){2-3} \cmidrule(lr){4-5} \cmidrule(lr){6-7} \cmidrule(lr){8-9} \cmidrule(lr){10-11} \cmidrule(lr){12-13} & {EM}                                                                  & {ES}                                       & {EM}                                        & {ES}                                        & {EM}                                       & {ES}                                       & {EM}      & {ES}      & {EM}      & {ES}      & {EM}      & {ES}      \\
      \midrule
      \textbf{Random-All}                                                                                                   & 6.18                                                                  & 46.19                                      & 33.94                                       & 70.98                                       & 28.32                                      & 66.87                                      & 31.45     & 69.09     & 26.93     & 62.13     & 36.69     & 74.42     \\
      \midrule
      \textbf{Top-p: 0.1}                                                                                                   & \bf 14.27                                                             & \bf 53.94                                  & 37.85                                       & 73.11                                       & 32.99                                      & \bf 68.75                                  & 34.16     & 70.43     & 29.19     & 62.09     & 40.98     & 76.26     \\
      \cmidrule(lr){2-3} \cmidrule(lr){4-5} \cmidrule(lr){6-7} \cmidrule(lr){8-9} \cmidrule(lr){10-11} \cmidrule(lr){12-13}
      \textbf{Top-p: 0.2}                                                                                                   & 13.52                                                                 & 53.20                                      & 38.04                                       & 73.13                                       & \bf 33.15                                  & 68.59                                      & 34.84     & 70.40     & 29.72     & 62.32     & 40.94     & \bf 76.25 \\
      \cmidrule(lr){2-3} \cmidrule(lr){4-5} \cmidrule(lr){6-7} \cmidrule(lr){8-9} \cmidrule(lr){10-11} \cmidrule(lr){12-13}
      \textbf{Top-p: 0.3}                                                                                                   & 12.88                                                                 & 52.60                                      & \bf 38.49                                   & \bf 73.19                                   & 32.84                                      & 68.31                                      & \bf 35.22 & \bf 70.64 & \bf 30.13 & \bf 62.77 & \bf 41.21 & 76.20     \\
      \bottomrule
    \end{tabular}
  }
  \caption{
    The results of completion using hierarchical context pruning with different top-p values.
  }
  \label{tab:hierarchical_topp_context_result_short}
\end{table*}

\paragraph{Current File.}
For the current file, any content within the file may be needed during completion, so we retain all content of the file and convert it into the Fill-in-the-middle (FIM) format.

\paragraph{Dependency Files.}
According to the experimental results in Section \ref{sec:cross_file_content_analysis}, removing the global context across files does not affect the accuracy of completions.
Therefore, for dependency files, we remove all global context from these files.

\paragraph{Other Files.}
We refer to files other than the current file and its direct dependency files, namely $\{F \setminus D_1(f)\} \setminus f\}$, collectively as other files.
For the content in \emph{other files}, we remove all global context, and then we employ \textbf{function-level} sampling and pruning methods to optimize the content of these files.

\begin{figure*}[h]
  \centering
  \includegraphics[width=0.98\textwidth]{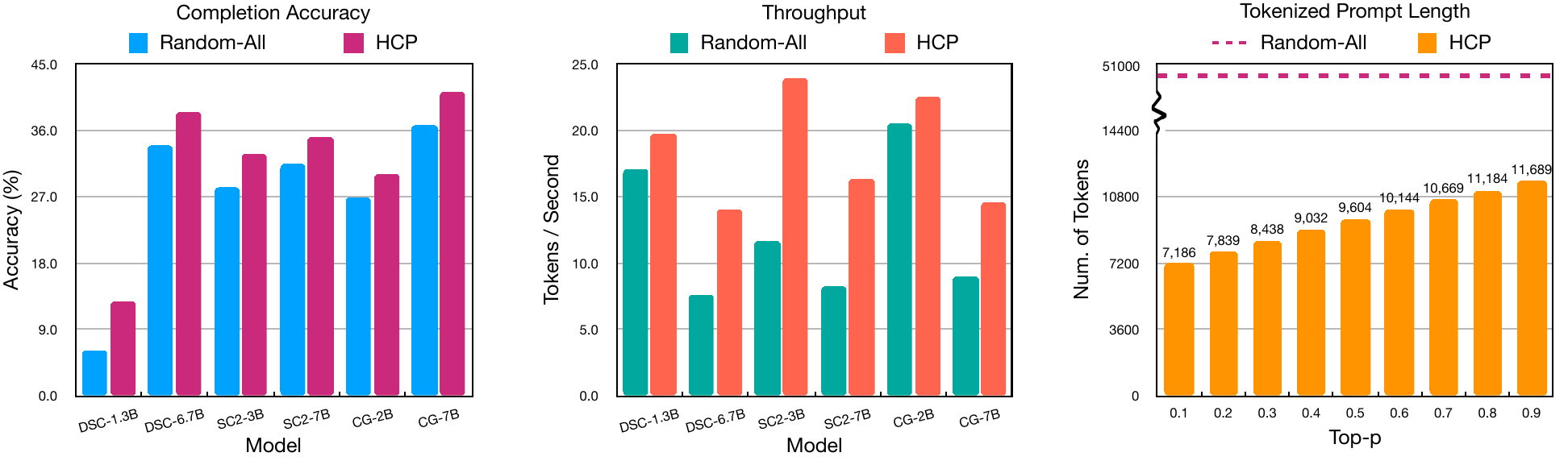}
  \caption{
    \textbf{left}: Comparison of completion results using random-all and the hierarchical context pruning across six models.
    \textbf{middle}: Comparison of throughput using random-all and the hierarchical context pruning across six models.
    \textbf{right}: Comparison of prompt length using random-all and the hierarchical context pruning of different top-p values (top-k=5).
  }
  \label{fig:hierarchical_results}
\end{figure*}

\subsection{Function-level Sampling}
\label{sec:function_level_sampling}
In this study, we used OpenAI's text-embedding API\footnote{openai-text-embedding-ada-002} to embed each function (or class method) and query code snippet in the repository. We then used the pre-computed similarity of embeddings between the query and candidate functions (or class methods) as an indicator of relevance.
We select the code from the current line of completion and the 10 lines before and after it as a query to find functions and class methods most relevant to the current completion content.

We implemented two sampling strategies (\textbf{top-k} and \textbf{top-p}) and designed distinct content pruning strategies for the functions (or class methods) sampled under each strategy, see Section \ref{sec:function_level_pruning}.

\subsection{Function-level Pruning}
\label{sec:function_level_pruning}
According to the experimental results in Section \ref{sec:cross_file_content_analysis}, the global context from all non-current files and most of the function bodies (or class method bodies) within the code repository can be pruned.
Appropriately pruning low-relevance content can significantly reduce the length of the prompt input to the model.

Let $F$ denote the set of all functions and class methods in the repository, $F_k$ represent the functions sampled using the top-k strategy, and $F_p$ represent the functions sampled using the top-p strategy:
\begin{equation}
  \begin{aligned}
    F_k & = \{f \mid f \in \text{Top}_k(F)\} \\
    F_p & = \{f \mid f \in \text{Top}_p(F)\} \\
  \end{aligned}
  \label{eq:function_sampling}
\end{equation}
where $F_k \subseteq F_p$.
Content from functions and class methods not within the set $F_k \cup F_p$ was completely pruned.
\paragraph{Top-k Context Pruning.}
For functions (or class methods) within the set $F_k$, we retained their entire content.
\paragraph{Top-p Context Pruning.}
For functions (or class methods) in the set $F_p$ but not in $F_k$, we prune their implementations and retained only their function headers (or class method headers).

\subsection{File-level Relevance Ranking}
Each function or class method in the repository has a similarity score.
We assign different relevance weights to functions sampled using different sampling strategies.
\begin{equation}
  \begin{aligned}
    W(f) & =
    \begin{cases}
      1.0, \quad \forall f \in F_k                        \\
      0.5, \quad \forall f \in F_p \setminus F_k          \\
      0.0, \quad \forall f \in F \setminus (F_k \cup F_p) \\
    \end{cases}
  \end{aligned}
  \label{eq:similarity_score}
\end{equation}
where $\text{Top}_k(F)$ and $\text{Top}_p(F)$ represent the functions with the highest relevance scores sampled using the top-k and top-p strategies, respectively.

The similarity of a class is defined as the weighted sum of its class methods:
\begin{equation}
  \begin{aligned}
     & S(c) = \sum_{m \in c} W(m) * S(m) \\
  \end{aligned}
  \label{eq:similarity_score_class}
\end{equation}
where, $c$ represents the class, and $m$ represents the class method.

The similarity of a file is defined as the weighted sum of its functions and classes:
\begin{equation}
  \begin{aligned}
     & S(f) = \sum_{x \in \mathcal{F}} W(x) * S(x) + \sum_{c \in \mathcal{C}} S(c) \\
  \end{aligned}
  \label{eq:similarity_score_file}
\end{equation}
where, $\mathcal{F}$ and $\mathcal{C}$ represent the set of functions and classes in the file, respectively.

Finally, we sort the files at the file-level according to the relevance score to determine their relative positions in the prompt.

\subsection{Experimental Results}
We initially fixed top-p at 1.0 and tested the impact of different top-k values on completion accuracy.
Table \ref{tab:hierarchical_topk_context_result_short} presents some of the experimental results, while Table \ref{tab:hierarchical_topk_context_result} in the Appendix \ref{sec:complete_sampling_results} provides a more comprehensive results.
We observed that increasing the top-k value beyond 5 did not result in significant improvements in accuracy.
Therefore, we conclude that a top-k value of 5 is sufficient.

We further fixed the top-k value at 5 and tested the impact of varying top-p values (ranging from 0.1 to 0.9) on completion accuracy.
Partial experimental results are presented in Table \ref{tab:hierarchical_topp_context_result_short}, with more comprehensive results available in Table \ref{tab:hierarchical_topp_context_result} in Appendix \ref{sec:complete_sampling_results}.
Our observations indicate that increasing the top-p value enhances completion accuracy;
however, beyond a top-p value of 0.3, the improvement in accuracy slows considerably.
Thus, we consider 0.3 to be a reasonable value.

Figure \ref{fig:hierarchical_results} visually compares the Hierarchical Context Pruning (HCP) strategy (top-k=5, top-p=0.3) with the method of randomly concatenating all repository code files across three dimensions: completion accuracy, throughput rate, and input length.
The visualization shows that, compared to random concatenation, \textbf{HCP} significantly reduces input length (enhancing throughput) while improving the model's completion accuracy.

\section{Conclusion}
In this study, we evaluated six Code LLMs pre-trained with repository-level code data.
We conducted a detailed error analysis on these Code LLMs, performed topological dependency analysis on files within the code repositories, and analyzed the content of these files.
Based on the results of these experiments, we proposed a strategy named Hierarchical Context Pruning to construct high-quality prompt inputs.
Finally, we conducted experiments on six Repo-Code LLMs to verify the effectiveness of the proposed method.

\section*{Limitations}
\paragraph{Benchmark.}
In this study, we utilized the CrossCodeEval benchmark for evaluation.
However, as demonstrated in the error analysis presented in Sections \ref{sec:error_analysis_main} and Appendix \ref{sec:error_analysis}, while the evaluation method based on exact matches is convenient and quick, it does not provide comprehensive results.
Therefore, there may be a discrepancy between the evaluation outcomes and the actual capabilities of the model.

\paragraph{Function-level Sampling.}
In this study, sampling functions and class methods based on relevance required the use of a text embedding model.
When the number of code files in the repository is excessive, this may reduce the sampling rate, leading to increased completion latency.

\section*{Ethical Statements}
This study does not involve human participants, personal data, or hazardous materials, and primarily focuses on computational model performance.
All resources used are open-source or properly licensed, ensuring compliance with relevant standards.

% \clearpage
\bibliography{acl2023}
\bibliographystyle{acl_natbib}

\appendix
\onecolumn
\section{Error Description and Analysis}
\label{sec:error_analysis}
In this section, we present detailed instances of various error types in model completions, accompanied by in-depth explanations and analyses of these errors.
Figures \ref{fig:error_redundant_content_generation}-\ref{fig:error_incorrect_content_generation} depict representatives for each error category.
Each figure is bifurcated, with the left panel showing the output generated by the code model and the right panel presenting the corresponding ground truth.
Errors in model completions are emphasized in \textcolor{red}{\textit{red italic}} text, whereas the ground truth is denoted in \textcolor{green}{\textit{green italic}}.

\subsection{Redundant Content Generation}
\begin{figure*}[h]
  \centering
  \includegraphics[width=\textwidth]{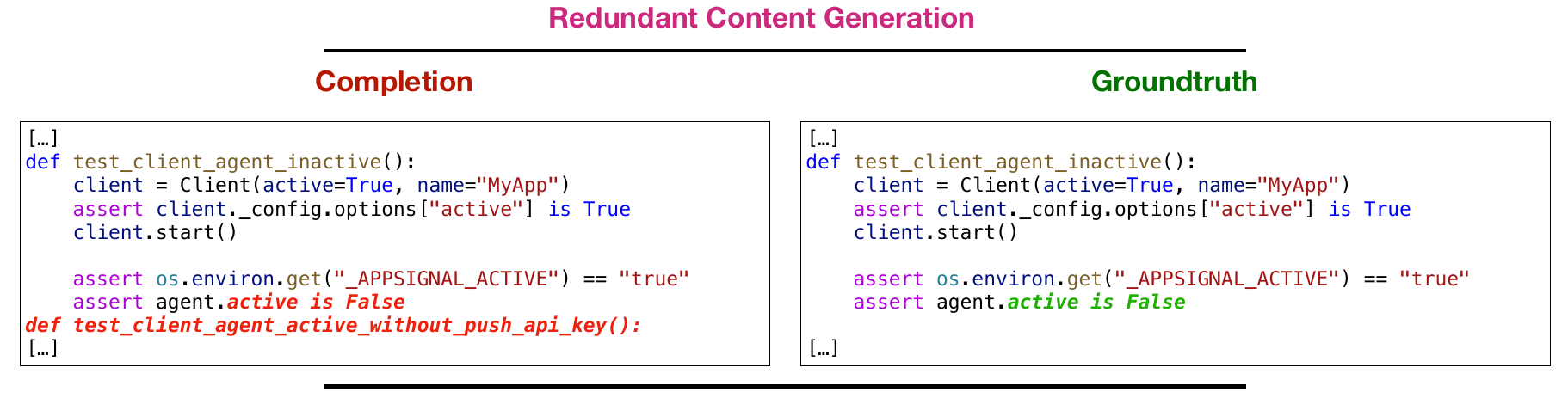}
  \caption{An example of redundant content generation error.}
  \label{fig:error_redundant_content_generation}
\end{figure*}

Redundant Content Generation means that the method is correctly called, but unnecessary additional content is generated.
Figure \ref{fig:error_redundant_content_generation} illustrates an example of a Redundant Content Generation error.
The ground truth specifies {\texttt{active is False}}, yet the model's completion includes not only \texttt{active is False} but also additional irrelevant content.

\subsection{Partial Content Missing}
\begin{figure*}[h]
  \centering
  \includegraphics[width=\textwidth]{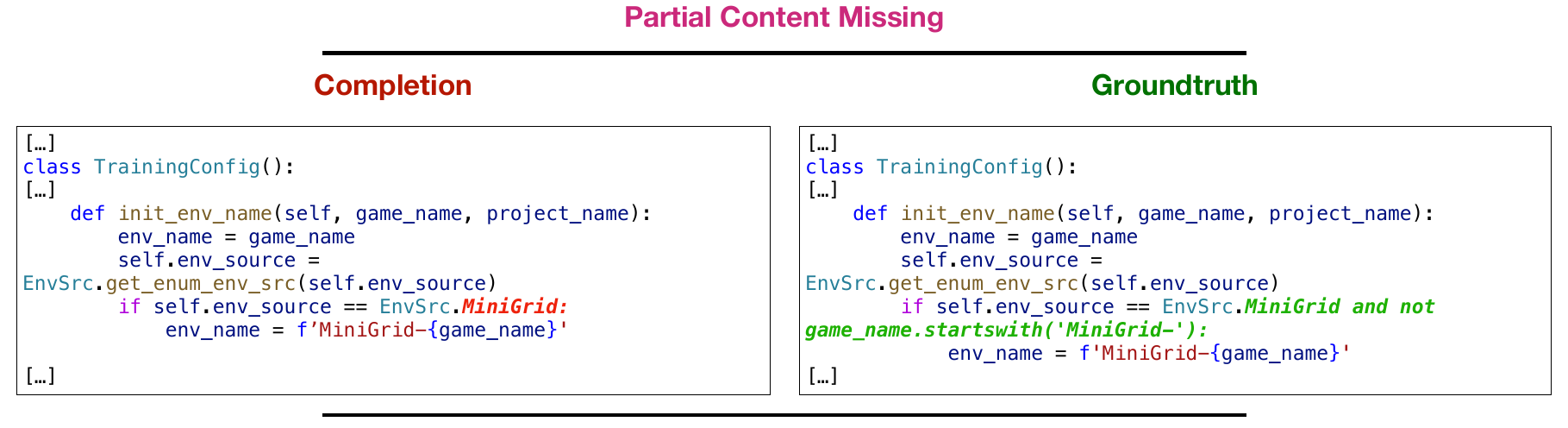}
  \caption{An example of partial content missing error.}
  \label{fig:error_partial_content_missing}
\end{figure*}

Partial Content Missing indicates that the right method is called, but the generated content is incomplete, although this might still be acceptable to the user.
Figure \ref{fig:error_partial_content_missing} presents an example of a Partial Content Missing error.
The ground truth is \texttt{MinGrid and not game\_name.startswith('MiniGrid-')}, but the code completion model only managed to replicate a portion of this ground truth.

\subsection{Parameter Value Error}
\begin{figure*}[h]
  \centering
  \includegraphics[width=\textwidth]{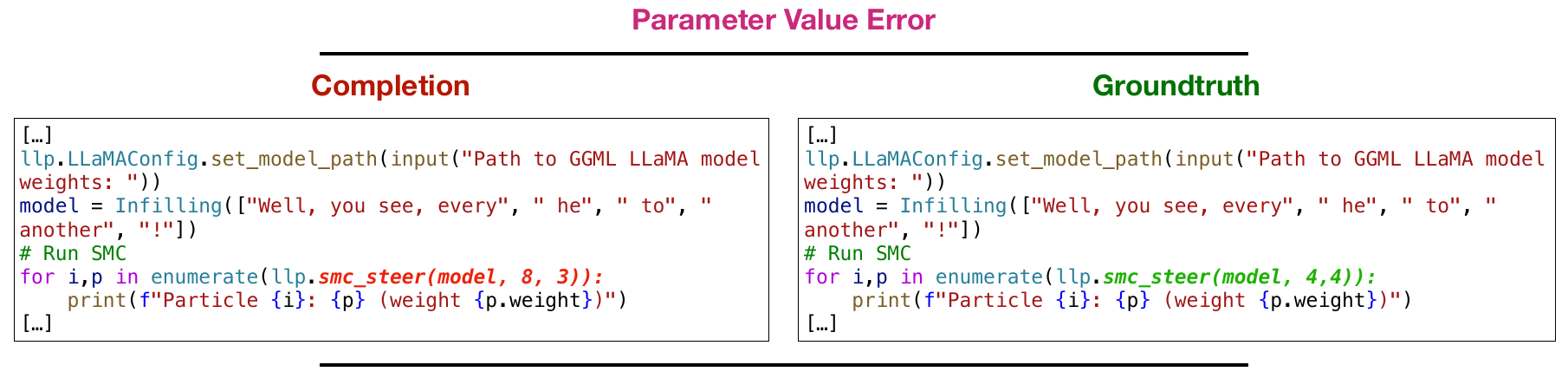}
  \caption{An example of parameter value error.}
  \label{fig:error_parameter_value_error}
\end{figure*}

The Parameter Value Error reflects the situation where the function call is correct, but the passed parameter values are incorrect.
Figure \ref{fig:error_parameter_value_error} displays an instance of a Parameter Value Error.
The code completion model correctly invokes the class method, but the parameters it employs differ from those specified in the ground truth.

\subsection{Exact Match Error}
\begin{figure*}[h]
  \centering
  \includegraphics[width=\textwidth]{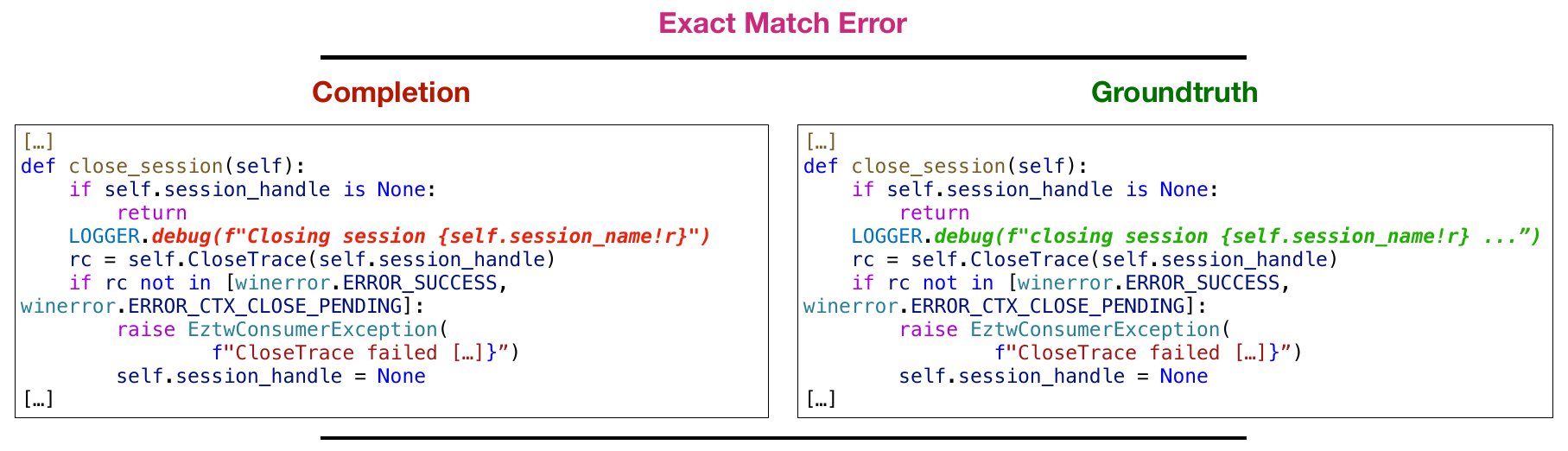}
  \caption{An example of exact match error.}
  \label{fig:error_exact_match_error}
\end{figure*}

Exact Match Error is a misjudgment due to the limitations of the exact match metric, such as using default values or specific strings when calling a function.
Figure \ref{fig:error_exact_match_error} illustrates an example of an Exact Match Error.
The content completed by the code model is syntactically correct and semantically accurate, differing only slightly in textual terms from the ground truth.
To avoid such misjudgments, a more reasonable evaluation method is necessary to assess the completion results.

\subsection{Non-existent Method Call}
\begin{figure*}[h]
  \centering
  \includegraphics[width=\textwidth]{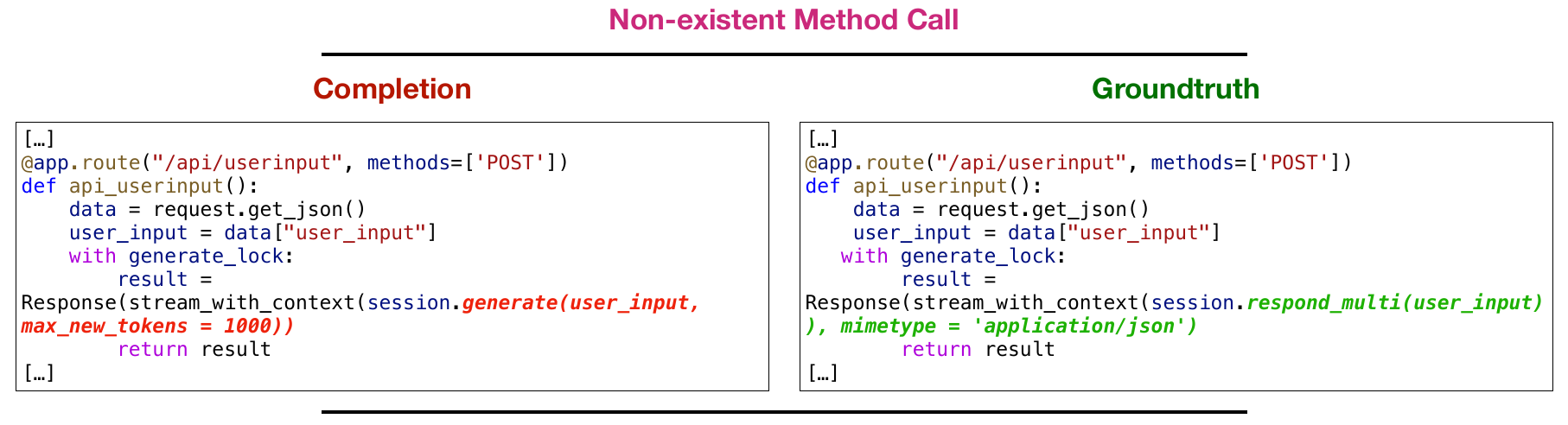}
  \caption{An example of non-existent method call error.}
  \label{fig:error_nonexistent_method_call}
\end{figure*}

Non-existent Method Call indicates a call to a function, method, or property that does not exist.
Figure \ref{fig:error_nonexistent_method_call} presents an example of a Non-existent Method Call error.
The ground truth refers to a class method within the \texttt{session} class; however, the content generated by the code completion model erroneously calls a method that does not exist in the \texttt{session} class.
This error can be regarded as a form of hallucination in the context of code completion.

\subsection{Improper Method Invocation}
\begin{figure*}[h]
  \centering
  \includegraphics[width=\textwidth]{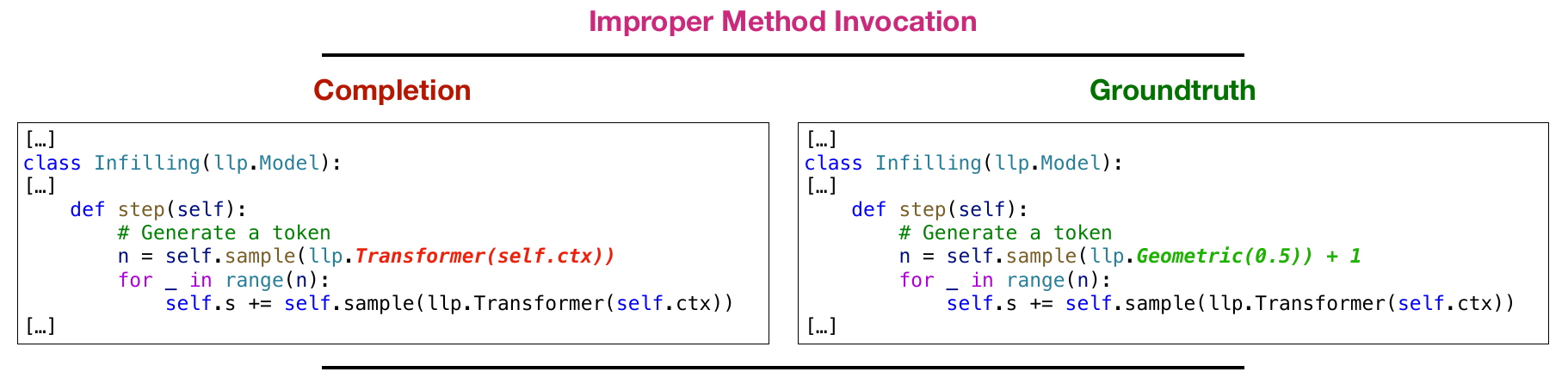}
  \caption{An example of improper method invocation error.}
  \label{fig:error_improper_method_invocation}
\end{figure*}

Improper Method Invocation represents the situation where the call is made to an existing method, but a different, more appropriate method should have been used.
Figure \ref{fig:error_improper_method_invocation} showcases an example of an Improper Method Invocation error.
The code completion model generated a call to the class method \texttt{Transformer} within the llp class, whereas the correct content should have invoked the class method \texttt{Geometric} within the same class.

\subsection{Missing Method Invocation}
\begin{figure*}[h]
  \centering
  \includegraphics[width=\textwidth]{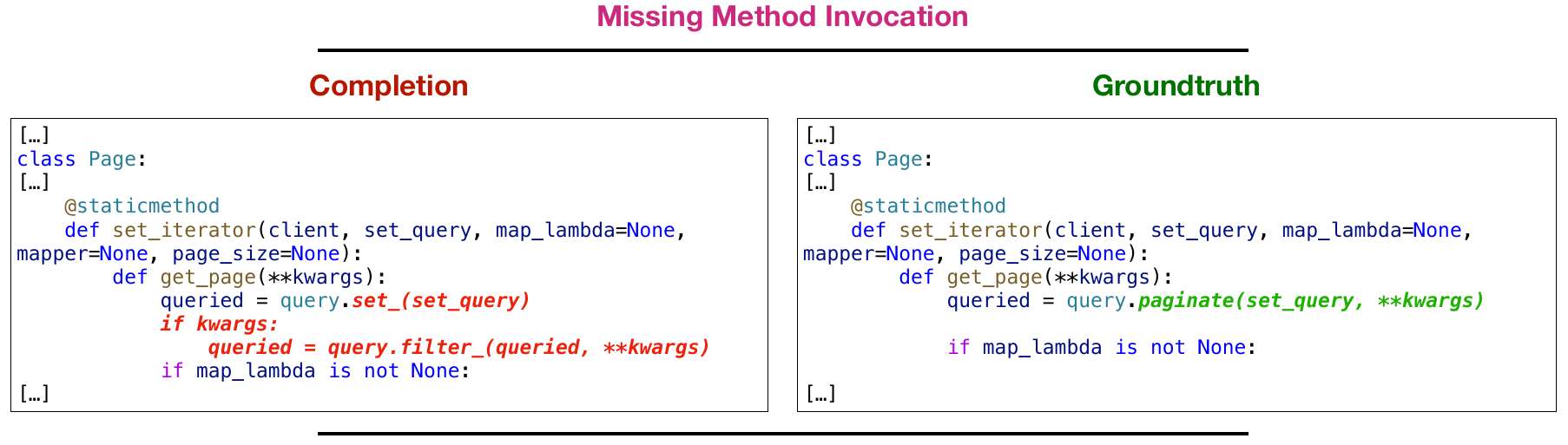}
  \caption{An example of missing method invocation error.}
  \label{fig:error_missing_method_invocation}
\end{figure*}

Missing Method Invocation indicates that a function or method should have been called to achieve functionality, but the model failed to make this call.
Figure \ref{fig:error_missing_method_invocation} illustrates an example of a Missing Method Invocation error.
The ground truth involves calling the class method \texttt{paginate} from the \texttt{query} class to obtain the \texttt{queried} variable.
However, the code completion model failed to complete this method invocation and instead achieved the same functionality through multiple alternative class methods.

\subsection{Incorrect Content Generation}
\begin{figure*}[h]
  \centering
  \includegraphics[width=\textwidth]{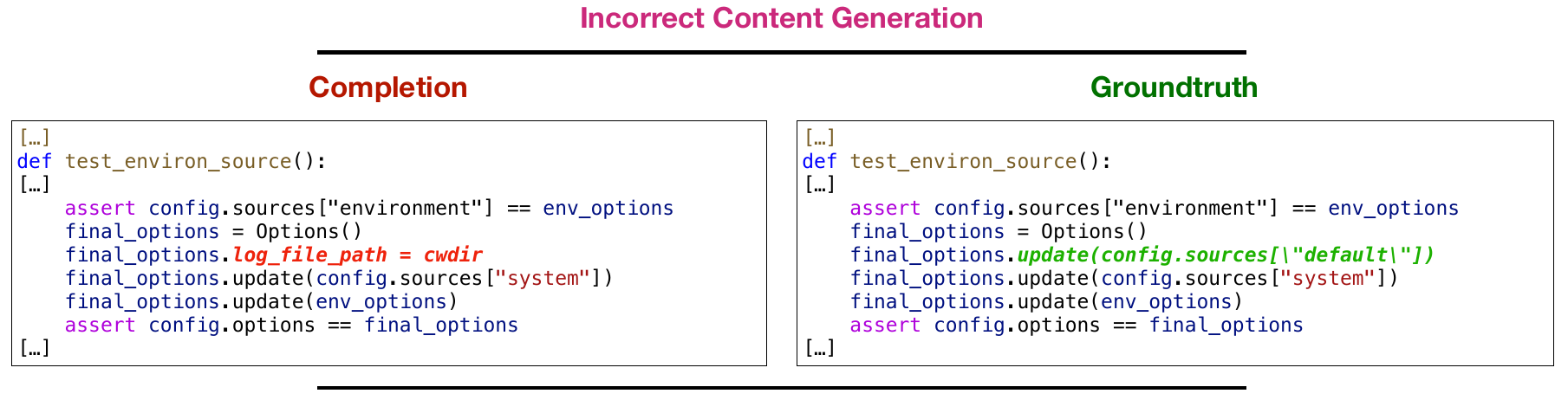}
  \caption{An example of incorrect content generation error.}
  \label{fig:error_incorrect_content_generation}
\end{figure*}

Incorrect Content Generation represents the situation where the generated content is illogical, irrelevant to the current code context, or completely incorrect.
Figure \ref{fig:error_incorrect_content_generation} depicts an example of an Incorrect Content Generation error. The content produced by the code completion model is entirely unrelated to the ground truth and also lacks relevance to the current code context.

\section{Special Tokens \& Prompt Templates}
Table \ref{tab:special_tokens} shows the special tokens used by DeepseekCoder, Starcoder2, and CodeGemma for fill-in-the-middle code completion.
The prompt templates for DeepseekCoder, Starcoder2, and CodeGemma are shown in Table \ref{tab:prompt_template}.
Both Starcoder2 and CodeGemma utilize special tokens for segmenting code files, whereas DeepseekCoder does not employ such tokens, despite being trained on repository-level code data.
\begin{table*}[!htbp]
  \centering
  \resizebox{\textwidth}{!}{
    \begin{tabular}{l l}
      \toprule
      \textbf{Model}                 & \textbf{\qquad\qquad\qquad\qquad\qquad\qquad Special Tokens}                           \\
      \midrule
      \multirow{1}{*}{DeepseekCoder} & \texttt{<|fim\_begin|>,<|fim\_hole|>,<|fim\_end|>}                                     \\
      \midrule
      \multirow{1}{*}{Starcoder2}    & \texttt{<repo\_name>,<file\_sep>,<fim\_pad>,<fim\_prefix>,<fim\_suffix>,<fim\_middle>} \\
      \midrule
      \multirow{1}{*}{CodeGemma}     & \texttt{<|file\_separator|>,<|fim\_prefix|>,<|fim\_suffix|>,<|fim\_middle|>}           \\
      \bottomrule
    \end{tabular}
  }
  \caption{Special tokens used by DeepseekCoder, Starcoder2 and CodeGemma for fill-in-the-middle code completion.}
  \label{tab:special_tokens}
\end{table*}

\begin{table*}[h]
  \centering
  \resizebox{\textwidth}{!}{
    \begin{tabular}{l l}
      \toprule
      \textbf{Model}                 & \textbf{\qquad\qquad\qquad\qquad\qquad Fill-in-the-Middle Prompt Template}                                                                                                                   \\
      \midrule
      \multirow{2}{*}{DeepseekCoder} & \texttt{\#file\_path0\textbackslash ncode0\textbackslash n\#file\_path1\textbackslash ncode1\textbackslash n\#file\_path2\textbackslash ncode2\textbackslash n\#file\_path3\textbackslash n} \\
                                     & \texttt{\textbf{<|fim\_begin|>}prefix\_code\textbf{<|fim\_hole|>}suffix\_code\textbf{<|fim\_end|>}}                                                                                          \\
      \midrule
      \multirow{2}{*}{Starcoder2}    & \texttt{\textbf{<repo\_name>}reponame\textbf{<file\_sep>}file\_path0\textbackslash ncode0\textbf{<file\_sep>}file\_path1}                                                                    \\
                                     & \texttt{\textbf{<fim\_prefix>}prefix\_code\textbf{<fim\_suffix>}suffix\_code\textbf{<fim\_middle>}}                                                                                          \\
      \midrule
      \multirow{2}{*}{CodeGemma}     & \texttt{\textbf{<|file\_separator|>}file\_path0\textbackslash ncode0\textbf{<file\_separator>}file\_path1\textbackslash n}                                                                   \\
                                     & \texttt{\textbf{<|fim\_prefix|>}prefix\_code\textbf{<|fim\_suffix|>}suffix\_code\textbf{<|fim\_middle|>}}                                                                                    \\
      \bottomrule
    \end{tabular}
  }
  \caption{Prompt templates for DeepseekCoder, Starcoder2 and CodeGemma.}
  \label{tab:prompt_template}
\end{table*}

\section{Prompt Length Distribution}
\begin{figure*}[h]
  \centering
  \includegraphics[width=\textwidth]{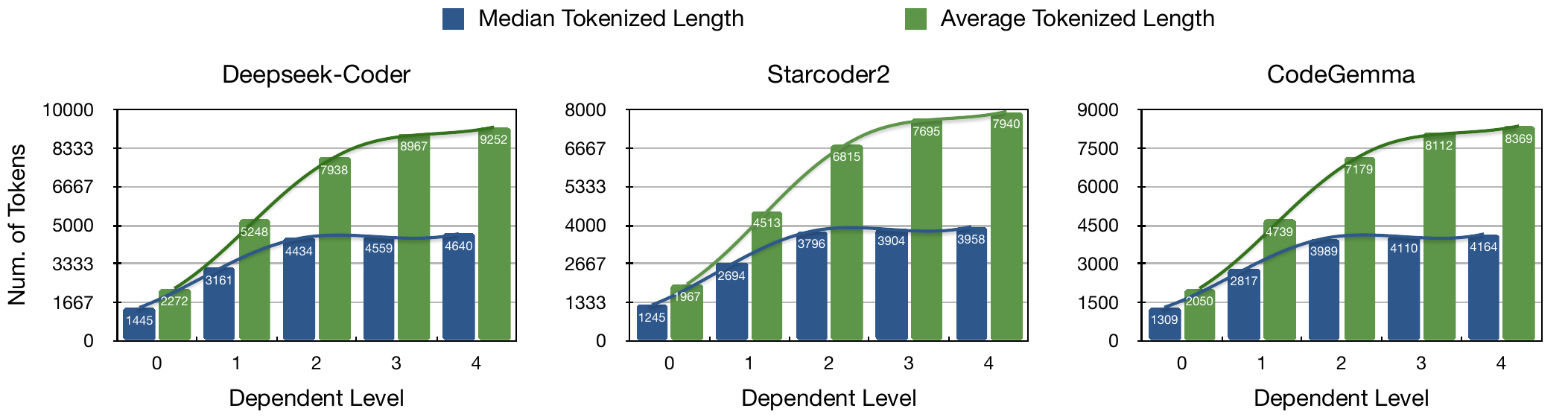}
  \caption{The distribution of tokenized prompt lengths in the CrossCodeEval benchmark. The x-aixs represents the dependent level, and the y-axis represents the number of tokens.
    \tikz[baseline=+0.1ex]\fill[NavyBlue] (0,0) rectangle (1em,0.7em); denotes the median value of the tokenized prompt length.
    \tikz[baseline=+0.1ex]\fill[OliveGreen] (0,0) rectangle (1em,0.7em); denotes the average value of the tokenized prompt length.}
  \label{fig:tokenized_prompt_length_distribution}
\end{figure*}

Table \ref{tab:num_tokens_over_dependent_level} presents the average and median lengths of input sequences for three code completion models when utilizing contexts of varying dependency levels.
Notably, Level $\infty$, which incorporates the entire repository code into the input, results in an average input sequence length exceeding 50,000, far surpassing the context window supported by these models.
To more visually observe the changes in input sequence length with respect to dependency levels, Figure \ref{fig:tokenized_prompt_length_distribution} was created.
It is evident that the median input sequence length begins to converge once the dependency level reaches 2, and the average input sequence length also starts to stabilize after reaching a dependency level of 3.

\begin{table*}[!htbp]
  \setlength\dashlinedash{1pt}
  \setlength\dashlinegap{1pt}
  \centering
  \resizebox{\textwidth}{!}{
    \begin{tabular}{l r r r r r r r r r r r r}
      \toprule
      \multirow{3}{*}{\bf Model}                                                                                          & \multicolumn{12}{c}{\textbf{CrossCodeEval Benchmark: Python}}                                                                                                                                \\\cmidrule(lr){2-13} &\multicolumn{2}{c}{\textbf{Level 0 }}&\multicolumn{2}{c}{\textbf{Level 1 }}&\multicolumn{2}{c}{\textbf{Level 2 }}&\multicolumn{2}{c}{\textbf{Level 3 }}&\multicolumn{2}{c}{\textbf{Level 4 }}&\multicolumn{2}{c}{\textbf{Level $\infty$ }}\\
      \cmidrule(lr){2-3} \cmidrule(lr){4-5} \cmidrule(lr){6-7} \cmidrule(lr){8-9}\cmidrule(lr){10-11}\cmidrule(lr){12-13} & {Median}                                                      & {Average} & {Median} & {Average} & {Median} & {Average} & {Median} & {Average} & {Median} & {Average} & {Median} & {Average} \\
      \midrule
      \textbf{DeepseekCoder}                                                                                              & 1,445                                                         & 2,272     & 3,161    & 5,248     & 4,434    & 7,938     & 4,559    & 8,967     & 4,640    & 9,252     & 44,475   & 58,217    \\
      \midrule
      \textbf{Starcoder2}                                                                                                 & 1,245                                                         & 1,967     & 2,694    & 4,513     & 3,796    & 6,815     & 3,904    & 7,695     & 3,958    & 7,940     & 38,174   & 50,632    \\
      \midrule
      \textbf{CodeGemma}                                                                                                  & 1,309                                                         & 2,050     & 2,817    & 4,739     & 3,989    & 7,179     & 4,110    & 8,112     & 4,164    & 8,369     & 39,647   & 52,875    \\
      \bottomrule
    \end{tabular}
  }
  \caption{
    The median and average tokenized prompt lengths of the DeepseekCoder, Starcoder2 and CodeGemma models on the CrossCodeEval: Python benchmark.
  }
  \label{tab:num_tokens_over_dependent_level}
\end{table*}

\section{Dependency Level Analysis}
\subsection{Complete Experimental Results}
Table \ref{tab:num_tokens_over_dependent_level} documents the comprehensive experimental results of repository file dependency analyses across six code completion models.
It is observed that when the length of the input sequence exceeds the model's context window, there is a significant decrease in completion accuracy.
However, truncating the input sequence from the left to fit within the model’s context window size reveals that greater amounts of code repository content can enhance completion accuracy.
Additionally, it was found that the DeepseekCoder-1.3B model exhibits a severe performance degradation in completion accuracy as the number of repository files increases.

\begin{table*}[h]
  \setlength\dashlinedash{1pt}
  \setlength\dashlinegap{1pt}
  \centering
  \resizebox{\textwidth}{!}{
    \begin{tabular}{l r r r r r r r r r r r r}
      \toprule
      \multirow{3}{*}{\bf Dependency}                                                                                       & \multicolumn{12}{c}{\textbf{Topological Dependency Analysis}}                                                                                                                                                                                                                                                                                                                                          \\
      \cmidrule(lr){2-13}                                                                                                   & \multicolumn{2}{c}{\textbf{DScoder-1.3B }}                    & \multicolumn{2}{c}{\textbf{DScoder-6.7B }} & \multicolumn{2}{c}{\textbf{Starcoder2-3B }} & \multicolumn{2}{c}{\textbf{Starcoder2-7B }} & \multicolumn{2}{c}{\textbf{CodeGemma-2B }} & \multicolumn{2}{c}{\textbf{CodeGemma-7B }}                                                                                                       \\
      \cmidrule(lr){2-3} \cmidrule(lr){4-5} \cmidrule(lr){6-7} \cmidrule(lr){8-9} \cmidrule(lr){10-11} \cmidrule(lr){12-13} & {EM}                                                          & {ES}                                       & {EM}                                        & {ES}                                        & {EM}                                       & {ES}                                       & {EM}           & {ES}           & {EM}           & {ES}           & {EM}           & {ES}           \\
      \midrule
      \textbf{Dep-Level: 0}                                                                                                 & \textbf{16.72}                                                & \textbf{56.60}                             & 28.14                                       & 68.40                                       & 21.92                                      & 61.45                                      & 23.16          & 63.62          & 20.60          & 55.97          & 30.40          & 69.76          \\
      \hdashline\noalign{\vskip 0.4ex}
      \; + {left truncate}                                                                                                  & 16.72                                                         & 56.58                                      & 28.14                                       & 68.36                                       & 21.92                                      & 61.49                                      & 22.98          & 63.58          & 20.64          & 56.26          & 30.58          & 70.36          \\
      \midrule
      \textbf{Dep-Level: 1}                                                                                                 & 14.99                                                         & 54.33                                      & 32.20                                       & 68.57                                       & 26.33                                      & 64.54                                      & 28.66          & 67.00          & 23.16          & 55.00          & 32.17          & 65.77          \\
      \hdashline\noalign{\vskip 0.4ex}
      \; + {left truncate}                                                                                                  & 15.44                                                         & 55.03                                      & 33.03                                       & 70.77                                       & 26.18                                      & 64.15                                      & 28.51          & 66.91          & 24.37          & 58.79          & 34.65          & 73.01          \\
      \midrule
      \textbf{Dep-Level: 2}                                                                                                 & 12.73                                                         & 51.72                                      & 30.21                                       & 65.46                                       & 26.63                                      & 64.50                                      & 29.83          & 67.03          & 21.24          & 49.62          & 28.36          & 57.76          \\
      \hdashline\noalign{\vskip 0.4ex}
      \; + {left truncate}                                                                                                  & 13.63                                                         & 53.45                                      & 33.56                                       & 70.74                                       & 26.70                                      & 64.58                                      & 29.45          & 67.03          & 25.31          & 59.27          & 35.67          & 73.26          \\
      \midrule
      \textbf{Dep-Level: 3}                                                                                                 & 12.28                                                         & 50.90                                      & 28.93                                       & 63.67                                       & 26.74                                      & 64.52                                      & 29.42          & 66.58          & 20.30          & 47.64          & 27.16          & 55.66          \\
      \hdashline\noalign{\vskip 0.4ex}
      \; + {left truncate}                                                                                                  & 13.26                                                         & 53.17                                      & 33.07                                       & 70.51                                       & 26.82                                      & 64.56                                      & 29.23          & 67.01          & 25.35          & 59.30          & 35.93          & 73.34          \\
      \midrule
      \textbf{Dep-Level: 4}                                                                                                 & 12.13                                                         & 50.69                                      & 28.44                                       & 63.15                                       & 26.48                                      & 64.30                                      & 29.68          & 66.84          & 20.08          & 47.29          & 26.93          & 55.16          \\
      \hdashline\noalign{\vskip 0.4ex}
      \; + {left truncate}                                                                                                  & 13.37                                                         & 53.20                                      & 33.22                                       & 70.57                                       & 26.59                                      & 64.46                                      & 29.53          & 67.07          & 25.54          & 59.42          & 36.12          & 73.54          \\
      \midrule
      \textbf{Dep-Level: $\infty$}                                                                                          & 1.32                                                          & 28.04                                      & 7.08                                        & 17.53                                       & 18.19                                      & 51.92                                      & 24.52          & 54.73          & 1.54           & 6.17           & 1.85           & 3.88           \\
      \hdashline\noalign{\vskip 0.4ex}
      \; + {left truncate}                                                                                                  & 5.76                                                          & 46.22                                      & \textbf{35.29}                              & \textbf{71.51}                              & \textbf{30.43}                             & \textbf{67.34}                             & \textbf{33.03} & \textbf{69.57} & \textbf{29.08} & \textbf{62.91} & \textbf{39.32} & \textbf{75.35} \\
      \bottomrule
    \end{tabular}
  }
  \caption{
    Comparison of completion results using different context dependency levels across 6 models.
    \textbf{EM} denotes Exact Match, and \textbf{ES} denotes Edit Similarity.
    \emph{$\infty$} denotes the prompt including all files in the repository.
    \emph{+left truncate} denotes the prompt is truncated to the max context window of LLMs from the left.
  }
  \label{tab:cross_file_result_full}
\end{table*}

\subsection{Hit Count Changes}
Table \ref{tab:hit_count_change} collates the variations in correct and incorrect completions across six code completion models when input contexts of different dependency levels are used.
It is evident that as the dependency level increases, the variations in the model's completion results become more stable.
This stability arises because the changes in the model's input context diminish as the dependency level is elevated.
This also indicates that augmenting the model’s input with additional content can enhance completion accuracy, albeit at the risk of turning some originally correct completions into incorrect ones.

We also observed that the DeepseekCoder series of models lack special tokens for delineating repository files; however, this deficiency does not result in more pronounced fluctuations in the outcomes.
This suggests that the DeepseekCoder models are capable of effectively distinguishing between different files in the repository, even without the aid of special tokens.
\begin{table*}[h]
  \centering
  \resizebox{\textwidth}{!}{
    \begin{tabular}{l c c c c c c}
      \toprule
      \multirow{2}{*}{\bf XF-Context}      & \multicolumn{6}{c}{\textbf{Hit Count Changes}}                                                                                                                                                                                                                                                                                                                                                 \\
      \cmidrule(lr){2-7}                   & \multicolumn{1}{c}{\textbf{DScoder-1.3B }}                    & \multicolumn{1}{c}{\textbf{DScoder-6.7B }}                     & \multicolumn{1}{c}{\textbf{Starcoder2-3B }}                   & \multicolumn{1}{c}{\textbf{Starcoder2-7B }}                   & \multicolumn{1}{c}{\textbf{CodeGemma-2B }}                    & \multicolumn{1}{c}{\textbf{CodeGemma-7B }}                    \\
      \cmidrule(lr){1-1}\cmidrule(lr){2-2}\cmidrule(lr){3-3}\cmidrule(lr){4-4}\cmidrule(lr){5-5}\cmidrule(lr){6-6}\cmidrule(lr){7-7}
      \textbf{Infile-Only}                 & \textbf{+444}                                                 & \textbf{+747}                                                  & \textbf{+582}                                                 & \textbf{+610}                                                 & \textbf{+548}                                                 & \textbf{+812}                                                 \\
      \cmidrule(lr){1-1}\cmidrule(lr){2-2}\cmidrule(lr){3-3}\cmidrule(lr){4-4}\cmidrule(lr){5-5}\cmidrule(lr){6-6}\cmidrule(lr){7-7}
      \textbf{\quad 0 $\rightarrow$ 1}     & \textbf{\textcolor{OliveGreen}{-108}\;\;\textcolor{red}{+74}} & \textbf{\textcolor{OliveGreen}{-47}\;\;\textcolor{red}{+177}}  & \textbf{\textcolor{OliveGreen}{-44}\;\;\textcolor{red}{+157}} & \textbf{\textcolor{OliveGreen}{-37}\;\;\textcolor{red}{+184}} & \textbf{\textcolor{OliveGreen}{-31}\;\;\textcolor{red}{+130}} & \textbf{\textcolor{OliveGreen}{-68}\;\;\textcolor{red}{+176}} \\
      \textbf{Level: 1}                    & \textbf{+408}                                                 & \textbf{+877}                                                  & \textbf{+695}                                                 & \textbf{+755}                                                 & \textbf{+647}                                                 & \textbf{+920}                                                 \\
      \cmidrule(lr){1-1}\cmidrule(lr){2-2}\cmidrule(lr){3-3}\cmidrule(lr){4-4}\cmidrule(lr){5-5}\cmidrule(lr){6-6}\cmidrule(lr){7-7}
      \textbf{\quad 1 $\rightarrow$ 2}     & \textbf{\textcolor{OliveGreen}{-61}\;\;\textcolor{red}{+13}}  & \textbf{\textcolor{OliveGreen}{-33}\;\;\textcolor{red}{+47}}   & \textbf{\textcolor{OliveGreen}{-41}\;\;\textcolor{red}{+55}}  & \textbf{\textcolor{OliveGreen}{-33}\;\;\textcolor{red}{+58}}  & \textbf{\textcolor{OliveGreen}{-30}\;\;\textcolor{red}{+55}}  & \textbf{\textcolor{OliveGreen}{-44}\;\;\textcolor{red}{+71}}  \\
      \textbf{Level: 2}                    & \textbf{+362}                                                 & \textbf{+891}                                                  & \textbf{+709}                                                 & \textbf{+782}                                                 & \textbf{+672}                                                 & \textbf{+947}                                                 \\
      \cmidrule(lr){1-1}\cmidrule(lr){2-2}\cmidrule(lr){3-3}\cmidrule(lr){4-4}\cmidrule(lr){5-5}\cmidrule(lr){6-6}\cmidrule(lr){7-7}
      \textbf{\quad 2 $\rightarrow$ 3}     & \textbf{\textcolor{OliveGreen}{-15}\;\;\textcolor{red}{+5}}   & \textbf{\textcolor{OliveGreen}{-20}\;\;\textcolor{red}{+7}}    & \textbf{\textcolor{OliveGreen}{-13}\;\;\textcolor{red}{+16}}  & \textbf{\textcolor{OliveGreen}{-19}\;\;\textcolor{red}{+13}}  & \textbf{\textcolor{OliveGreen}{-10}\;\;\textcolor{red}{+11}}  & \textbf{\textcolor{OliveGreen}{-11}\;\;\textcolor{red}{+18}}  \\
      \textbf{Level: 3}                    & \textbf{+352}                                                 & \textbf{+878}                                                  & \textbf{+712}                                                 & \textbf{+776}                                                 & \textbf{+673}                                                 & \textbf{+954}                                                 \\
      \cmidrule(lr){1-1}\cmidrule(lr){2-2}\cmidrule(lr){3-3}\cmidrule(lr){4-4}\cmidrule(lr){5-5}\cmidrule(lr){6-6}\cmidrule(lr){7-7}
      \textbf{\quad 3 $\rightarrow$ 4}     & \textbf{\textcolor{OliveGreen}{-3}\;\;\textcolor{red}{+6}}    & \textbf{\textcolor{OliveGreen}{-1}\;\;\textcolor{red}{+5}}     & \textbf{\textcolor{OliveGreen}{-10}\;\;\textcolor{red}{+4}}   & \textbf{\textcolor{OliveGreen}{-3}\;\;\textcolor{red}{+11}}   & \textbf{\textcolor{OliveGreen}{-3}\;\;\textcolor{red}{+8}}    & \textbf{\textcolor{OliveGreen}{-5}\;\;\textcolor{red}{+10}}   \\
      \textbf{Level: 4}                    & \textbf{+355}                                                 & \textbf{+882}                                                  & \textbf{+706}                                                 & \textbf{+784}                                                 & \textbf{+678}                                                 & \textbf{+959}                                                 \\
      \cmidrule(lr){1-1}\cmidrule(lr){2-2}\cmidrule(lr){3-3}\cmidrule(lr){4-4}\cmidrule(lr){5-5}\cmidrule(lr){6-6}\cmidrule(lr){7-7}
      \textbf{\; 4 $\rightarrow$ $\infty$} & \textbf{\textcolor{OliveGreen}{-238}\;\;\textcolor{red}{+36}} & \textbf{\textcolor{OliveGreen}{-135}\;\;\textcolor{red}{+190}} & \textbf{\textcolor{OliveGreen}{-55}\;\;\textcolor{red}{+157}} & \textbf{\textcolor{OliveGreen}{-68}\;\;\textcolor{red}{+161}} & \textbf{\textcolor{OliveGreen}{-45}\;\;\textcolor{red}{+139}} & \textbf{\textcolor{OliveGreen}{-72}\;\;\textcolor{red}{+157}} \\
      \textbf{Level: $\infty$}             & \textbf{+153}                                                 & \textbf{+937}                                                  & \textbf{+808}                                                 & \textbf{+877}                                                 & \textbf{+772}                                                 & \textbf{+1044}                                                \\
      \bottomrule
    \end{tabular}
  }
  \caption{
    The changes in the hit counts of correct and incorrect completions across six code completion models when using different context dependency levels.
    The green values denote the number of test samples that were originally correct but became incorrect as the dependency level of the input context increased.
    The red values represent the number of test samples that were initially incorrect but became correct with the elevation of the input context's dependency level.
  }
  \label{tab:hit_count_change}
\end{table*}

\begin{table*}[h]
  \setlength\dashlinedash{1pt}
  \setlength\dashlinegap{1pt}
  \centering
  \resizebox{\textwidth}{!}{
    \begin{tabular}{l r r r r r r r r r r r r}
      \toprule
      \multirow{3}{*}{\bf XF-Context}                                                                                       & \multicolumn{12}{c}{\textbf{Hierarchical Context Pruning (Top-p: 1.0)}}                                                                                                                                                                                                                                                                                                            \\
      \cmidrule(lr){2-13}                                                                                                   & \multicolumn{2}{c}{\textbf{DScoder-1.3B }}                              & \multicolumn{2}{c}{\textbf{DScoder-6.7B }} & \multicolumn{2}{c}{\textbf{Starcoder2-3B }} & \multicolumn{2}{c}{\textbf{Starcoder2-7B }} & \multicolumn{2}{c}{\textbf{CodeGemma-2B }} & \multicolumn{2}{c}{\textbf{CodeGemma-7B }}                                                                         \\
      \cmidrule(lr){2-3} \cmidrule(lr){4-5} \cmidrule(lr){6-7} \cmidrule(lr){8-9} \cmidrule(lr){10-11} \cmidrule(lr){12-13} & {EM}                                                                    & {ES}                                       & {EM}                                        & {ES}                                        & {EM}                                       & {ES}                                       & {EM}      & {ES}      & {EM}      & {ES}      & {EM}      & {ES}      \\
      \midrule
      \textbf{Random-All}                                                                                                   & 6.18                                                                    & 46.19                                      & 33.94                                       & 70.98                                       & 28.32                                      & 66.87                                      & 31.45     & 69.09     & 26.93     & 62.13     & 36.69     & 74.42     \\
      \midrule
      \textbf{Top-k: 5}                                                                                                     & 9.64                                                                    & 49.78                                      & 39.74                                       & 73.90                                       & 32.68                                      & 69.05                                      & 35.76     & 71.41     & 31.26     & 63.74     & 42.44     & 76.95     \\
      \cmidrule(lr){2-3} \cmidrule(lr){4-5} \cmidrule(lr){6-7} \cmidrule(lr){8-9} \cmidrule(lr){10-11} \cmidrule(lr){12-13}
      \textbf{Top-k: 10}                                                                                                    & \bf 9.91                                                                & \bf 49.85                                  & 40.30                                       & 74.56                                       & 34.15                                      & 69.37                                      & 36.47     & 71.50     & 31.82     & 64.34     & 42.63     & 77.35     \\
      \cmidrule(lr){2-3} \cmidrule(lr){4-5} \cmidrule(lr){6-7} \cmidrule(lr){8-9} \cmidrule(lr){10-11} \cmidrule(lr){12-13}
      \textbf{Top-k: 15}                                                                                                    & 9.23                                                                    & 49.23                                      & 40.75                                       & \bf 74.59                                   & 33.96                                      & 69.41                                      & 36.62     & 71.59     & 31.86     & 64.53     & 42.55     & 77.06     \\
      \cmidrule(lr){2-3} \cmidrule(lr){4-5} \cmidrule(lr){6-7} \cmidrule(lr){8-9} \cmidrule(lr){10-11} \cmidrule(lr){12-13}
      \textbf{Top-k: 20}                                                                                                    & 9.01                                                                    & 48.95                                      & \bf 41.24                                   & 74.57                                       & \bf 34.37                                  & \bf 69.81                                  & 36.66     & 71.56     & 31.93     & 64.67     & 42.85     & \bf 77.52 \\
      \cmidrule(lr){2-3} \cmidrule(lr){4-5} \cmidrule(lr){6-7} \cmidrule(lr){8-9} \cmidrule(lr){10-11} \cmidrule(lr){12-13}
      \textbf{Top-k: 25}                                                                                                    & 8.93                                                                    & 48.82                                      & 40.34                                       & 74.47                                       & 33.73                                      & 69.62                                      & \bf 37.00 & \bf 71.91 & \bf 32.46 & \bf 64.81 & \bf 43.11 & 77.47     \\
      \cmidrule(lr){2-3} \cmidrule(lr){4-5} \cmidrule(lr){6-7} \cmidrule(lr){8-9} \cmidrule(lr){10-11} \cmidrule(lr){12-13}
      \textbf{Top-k: 30}                                                                                                    & 8.44                                                                    & 48.48                                      & 39.74                                       & 74.17                                       & 33.28                                      & 69.42                                      & 36.14     & 71.29     & 32.46     & 64.81     & 42.44     & 77.20     \\
      \bottomrule
    \end{tabular}
  }
  \caption{
    The results of completion using hierarchical context pruning with different top-k values.
  }
  \label{tab:hierarchical_topk_context_result}
\end{table*}

\begin{table*}[h]
  \setlength\dashlinedash{1pt}
  \setlength\dashlinegap{1pt}
  \centering
  \resizebox{\textwidth}{!}{
    \begin{tabular}{l r r r r r r r r r r r r}
      \toprule
      \multirow{3}{*}{\bf XF-Context}                                                                                       & \multicolumn{12}{c}{\textbf{Hierarchical Context Pruning (Top-k: 5)}}                                                                                                                                                                                                                                                                                    \\
      \cmidrule(lr){2-13}                                                                                                   & \multicolumn{2}{c}{\textbf{DScoder-1.3B }}                            & \multicolumn{2}{c}{\textbf{DScoder-6.7B }} & \multicolumn{2}{c}{\textbf{Starcoder2-3B }} & \multicolumn{2}{c}{\textbf{Starcoder2-7B }} & \multicolumn{2}{c}{\textbf{CodeGemma-2B }} & \multicolumn{2}{c}{\textbf{CodeGemma-7B }}                                                 \\
      \cmidrule(lr){2-3} \cmidrule(lr){4-5} \cmidrule(lr){6-7} \cmidrule(lr){8-9} \cmidrule(lr){10-11} \cmidrule(lr){12-13} & {EM}                                                                  & {ES}                                       & {EM}                                        & {ES}                                        & {EM}                                       & {ES}                                       & {EM}  & {ES}  & {EM}  & {ES}  & {EM}  & {ES}  \\
      \midrule
      \textbf{Random-All}                                                                                                   & 6.18                                                                  & 46.19                                      & 33.94                                       & 70.98                                       & 28.32                                      & 66.87                                      & 31.45 & 69.09 & 26.93 & 62.13 & 36.69 & 74.42 \\
      \midrule
      \textbf{Top-p: 0.1}                                                                                                   & 14.27                                                                 & 53.94                                      & 37.85                                       & 73.11                                       & 32.99                                      & 68.75                                      & 34.16 & 70.43 & 29.19 & 62.09 & 40.98 & 76.26 \\
      \cmidrule(lr){2-3} \cmidrule(lr){4-5} \cmidrule(lr){6-7} \cmidrule(lr){8-9} \cmidrule(lr){10-11} \cmidrule(lr){12-13}
      \textbf{Top-p: 0.2}                                                                                                   & 13.52                                                                 & 53.20                                      & 38.04                                       & 73.13                                       & 33.15                                      & 68.59                                      & 34.84 & 70.40 & 29.72 & 62.32 & 40.94 & 76.25 \\
      \cmidrule(lr){2-3} \cmidrule(lr){4-5} \cmidrule(lr){6-7} \cmidrule(lr){8-9} \cmidrule(lr){10-11} \cmidrule(lr){12-13}
      \textbf{Top-p: 0.3}                                                                                                   & 12.88                                                                 & 52.60                                      & 38.49                                       & 73.19                                       & 32.84                                      & 68.31                                      & 35.22 & 70.64 & 30.13 & 62.77 & 41.21 & 76.20 \\
      \cmidrule(lr){2-3} \cmidrule(lr){4-5} \cmidrule(lr){6-7} \cmidrule(lr){8-9} \cmidrule(lr){10-11} \cmidrule(lr){12-13}
      \textbf{Top-p: 0.4}                                                                                                   & 11.60                                                                 & 51.58                                      & 38.42                                       & 72.92                                       & 32.81                                      & 68.51                                      & 35.07 & 70.55 & 30.40 & 62.97 & 40.98 & 76.30 \\
      \cmidrule(lr){2-3} \cmidrule(lr){4-5} \cmidrule(lr){6-7} \cmidrule(lr){8-9} \cmidrule(lr){10-11} \cmidrule(lr){12-13}
      \textbf{Top-p: 0.5}                                                                                                   & 11.49                                                                 & 51.50                                      & 38.95                                       & 73.31                                       & 32.88                                      & 68.67                                      & 34.73 & 70.33 & 30.17 & 62.73 & 41.32 & 76.15 \\
      \cmidrule(lr){2-3} \cmidrule(lr){4-5} \cmidrule(lr){6-7} \cmidrule(lr){8-9} \cmidrule(lr){10-11} \cmidrule(lr){12-13}
      \textbf{Top-p: 0.6}                                                                                                   & 11.00                                                                 & 51.14                                      & 38.87                                       & 73.33                                       & 32.32                                      & 68.56                                      & 34.73 & 70.38 & 30.02 & 63.15 & 41.58 & 76.20 \\
      \cmidrule(lr){2-3} \cmidrule(lr){4-5} \cmidrule(lr){6-7} \cmidrule(lr){8-9} \cmidrule(lr){10-11} \cmidrule(lr){12-13}
      \textbf{Top-p: 0.7}                                                                                                   & 11.11                                                                 & 51.21                                      & 38.83                                       & 73.52                                       & 31.94                                      & 68.14                                      & 34.92 & 70.74 & 30.47 & 63.13 & 41.77 & 76.40 \\
      \cmidrule(lr){2-3} \cmidrule(lr){4-5} \cmidrule(lr){6-7} \cmidrule(lr){8-9} \cmidrule(lr){10-11} \cmidrule(lr){12-13}
      \textbf{Top-p: 0.8}                                                                                                   & 10.40                                                                 & 50.39                                      & 38.95                                       & 73.56                                       & 31.79                                      & 68.34                                      & 34.80 & 70.59 & 30.17 & 63.05 & 41.81 & 76.41 \\
      \cmidrule(lr){2-3} \cmidrule(lr){4-5} \cmidrule(lr){6-7} \cmidrule(lr){8-9} \cmidrule(lr){10-11} \cmidrule(lr){12-13}
      \textbf{Top-p: 0.9}                                                                                                   & 10.40                                                                 & 50.08                                      & 38.61                                       & 73.13                                       & 31.94                                      & 68.26                                      & 34.54 & 70.22 & 30.43 & 63.18 & 41.81 & 76.51 \\
      \bottomrule
    \end{tabular}
  }
  \caption{
    The results of completion using hierarchical context pruning with different top-p values.
  }
  \label{tab:hierarchical_topp_context_result}
\end{table*}

\section{Complete Funcation-Level Sampling Experiment Results}
\label{sec:complete_sampling_results}
Due to space constraints, we report only a subset of the results from the function-level sampling experiments in the main body.
Tables \ref{tab:hierarchical_topk_context_result} and \ref{tab:hierarchical_topp_context_result} provide a comprehensive statistical overview of the complete sampling experiments.
\subsection{Top-k Sampling}
Table \ref{tab:hierarchical_topk_context_result} details the results of top-k sampling, where top-p is fixed at 1.0.
It is observed that increasing the value of k does not significantly enhance the accuracy of completions; improvements become negligible when k exceeds 5.
\subsection{Top-p Sampling}
Table \ref{tab:hierarchical_topp_context_result}, on the other hand, presents the outcomes of top-p sampling with top-k fixed at 5.
Here, increasing the value of p does not yield significant improvements, particularly when p exceeds 0.3.

\section{Dependency Search Algorithm}
Algorithm \ref{alg:dependency-modeling} delineates the specific process we employed for dependency modeling within code repositories.
For more detailed implementation specifics, please visit our code repository.

\begin{algorithm*}
  \caption{Dependency Search Algorithm for Python Files}
  \label{alg:dependency-modeling}
  % \resizebox{\textwidth}{!}{
  \begin{algorithmic}[1] % The number [1] here indicates line numbers are shown
    \State \textbf{Input:} file - initial Python file, maxDepth - maximum search depth
    \State \textbf{Output:} List of dependent files

    \Function{FindDependencies}{file, maxDepth}
    \State queue = [(file, 0)]  % Initialize queue with the start file and depth 0
    \State visited = set()      % Set to track visited files

    \While{queue}
    \State currentFile, currentDepth = queue.pop(0)  % Dequeue the first element
    \If{currentDepth > maxDepth}
    \State \textbf{break}
    \EndIf
    \State imports = extractImports(currentFile)
    \For{imp in imports}
    \If{imp is local and imp not in visited}
    \State visited.add(imp)
    \State queue.append((imp, currentDepth + 1))
    \EndIf
    \EndFor
    \EndWhile
    \State \textbf{return} visited
    \EndFunction

    \Function{extractImports}{file}
    \State \textit{Use Tree-Sitter to parse file and extract all import statements}
    \State \textbf{return} list of imports
    \EndFunction
  \end{algorithmic}
  % }
\end{algorithm*}

\section{Additional Related Work}
\subsection{Long Context Code Large Language Models}
Research on optimizing large language models for long contexts has been underway for some time, with many innovative long-context optimization techniques
\citep{chen2023extending,chen2024longlora,xiao2024efficient,ding2024longrope,chen2024long}
and evaluation sets
\citep{bai2023longbench,zhang2023marathon,zhang2024inftybench}
being proposed and widely applied.
Some Code LLMs \citep{guo2024deepseekcoder,lozhkov2024starcoder} utilize these techniques for fine-tuning to extend their context windows.
Larger context windows allow Code LLMs to receive and process more complex code content, such as repository-level code completion and repository-level code repair \citep{ding2023crosscodeeval,liu2023repobench,Zhang_2024,li2024evocodebench}.

\subsection{Code Agent}
Research on Code Agents \citep{Yang2024SWEagentAI,fang2024llm,thakur2024incontext,shi2024ehragent} focuses on developing intelligent systems that assist in software development by automating tasks like code generation and debugging \citep{holt2024l2mac,wang2023executionbased,yin2022natural,zelikman2024selftaught} .
The use of Code LLMs has proven effective in understanding complex code structures and semantics \citep{wang2024executable,yang2023intercode}.
These models have been further refined to handle specific software development tasks, including repository-level code analysis and automated error correction \citep{jimenez2024swebench}.

\end{document}